\documentclass[conference]{IEEEtran}

\usepackage[usenames,dvipsnames,svgnames,table]{xcolor}
\usepackage{wrapfig}
\usepackage{times}
\usepackage[numbers]{natbib}
\usepackage{multicol}
\usepackage[bookmarks=true]{hyperref}
\usepackage{graphicx}
\usepackage{amsmath,amsthm,amssymb}
\usepackage{pgf,tikz}
\usepackage{mathrsfs}
\usetikzlibrary{arrows}
\usepackage[linesnumbered,ruled,vlined]{algorithm2e}
\usepackage{epstopdf}
\usepackage{overpic}
\usepackage{xcolor}
\usepackage{xargs} 
\usepackage{verbatim} 
\usepackage[textsize=footnotesize]{todonotes}
\usepackage{enumerate}
\usepackage{placeins} 
\usepackage{marvosym}

\usetikzlibrary{
    arrows,
    arrows.meta,
    shapes,
    shapes.geometric,
    decorations.markings}
\tikzset{every node/.style={thick,shape=circle,draw=black,fill=blue!15,text=black,minimum size=1.7em,inner sep=0.5}}
\tikzset{every picture/.style=thick}
\tikzset{>={Latex[length=2mm]}}
\colorlet{customLightRed}{red!65}     

\usepackage{etoolbox}
\apptocmd{\thebibliography}{\raggedright}{}{}

\let\oldsection\section
\renewcommand{\section}{
  \FloatBarrier\oldsection
}

\newcommand{\changed}[1]{{\color{black} #1}}

\newcommand{\Wspace}{{\mathcal{W}}}
\newcommand{\objects}{{\mathcal{O}}}
\newcommand{\actions}{{\mathcal{A}}}
\newcommand{\Cspace}{{\mathcal{C}}}
\newcommand{\Fspace}{{\mathcal{F}}}

\newcommand{\toro}{{\tt TORO}\xspace}

\newcommand{\rno}{{\tt TORO-NO}\xspace}
\newcommand{\uno}{{\tt TORO-UNO}\xspace}
\newcommand{\rwo}{{\tt TORO}\xspace}
\newcommand{\tsp}{{\tt TSP}\xspace}

\newcommand{\fvs}{{\tt FVS}\xspace}

\newcommand{\DAG}{DAG\xspace}

\makeatletter
\def\thm@space@setup{\thm@preskip=0pt
\thm@postskip=0pt}
\makeatother

\newcommandx{\nt}[2][1=]{\todo[linecolor=blue,
			backgroundcolor=blue!10,bordercolor=blue,#1]{#2}}

\newtheorem{problem}{Problem}
\newtheorem{lemma}{Lemma}[section]
\newtheorem{observation}{Observation}[section]

\newtheorem{theorem}{Theorem}[section]
\theoremstyle{definition}

\newtheorem{remark}{Remark}

{\end{list}}

\makeatletter 
\renewcommand\large{\@setfontsize\large{9pt}{18}}
\makeatother

\begin{document}

\title{\huge High-Quality Tabletop Rearrangement with Overhand Grasps:\\
Hardness Results and Fast Methods}

\author{Author Names Omitted for Anonymous Review. Paper-ID 171}


\author{%
    \authorblockN{
        Shuai D. Han,
        Nicholas M. Stiffler,
        Athansios Krontiris, 
        Kostas E. Bekris, and
        Jingjin Yu}
    \authorblockA{%
        Computer Science Department\\
        Rutgers, the State University of New Jersey,
        Piscataway, New Jersey, USA\\%
        Email: \{shuai.han, nick.stiffler, tdk.krontir, kostas.bekris, jingjin.yu\}\hspace*{.25em}\MVAt \hspace*{.25em}rutgers.edu%
    }
}

\maketitle

\begin{abstract}
This paper studies the underlying combinatorial structure of a class
of object rearrangement problems, which appear frequently in
applications. The problems involve multiple, similar-geometry objects
placed on a flat, horizontal surface, where a robot can approach them
from above and perform pick-and-place operations to rearrange them.
The paper considers both the case where the start and goal object
poses overlap, and where they do not.  For overlapping poses, the
primary objective is to minimize the number of pick-and-place actions
and then to minimize the distance traveled by the end-effector. For
the non-overlapping case, the objective is solely to minimize the
end-effector distance. While such problems do not involve all the
complexities of general rearrangement, they remain computationally
hard challenges in both cases.  This is shown through two-way
reductions between well-understood, hard combinatorial challenges and
these rearrangement problems. The benefit of the reduction is that
there are well studied algorithms for solving these well-established
combinatorial challenges.  These algorithms can be very efficient in
practice despite the hardness results. The paper builds on these
reduction results to propose an algorithmic pipeline for dealing with
the rearrangement problems.  Experimental evaluation shows that the
proposed pipeline achieves high-quality paths with regards to the
optimization objectives. Furthermore, it exhibits highly desirable
scalability as the number of objects increases in both the overlapping
and non-overlapping setups.
\end{abstract}

\IEEEpeerreviewmaketitle

\section{Introduction}
\label{sec:introduction}

\begin{figure}[t]
  \centering
   \includegraphics[width = 0.49 \columnwidth, height = 1.2in]{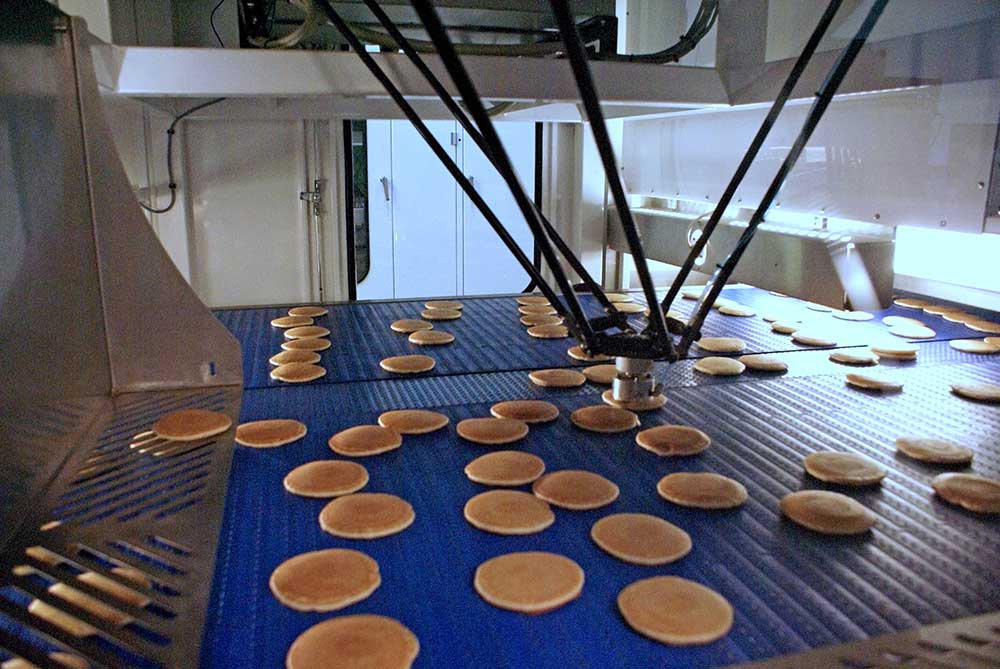}
   \includegraphics[width = 0.49 \columnwidth, height = 1.2in]{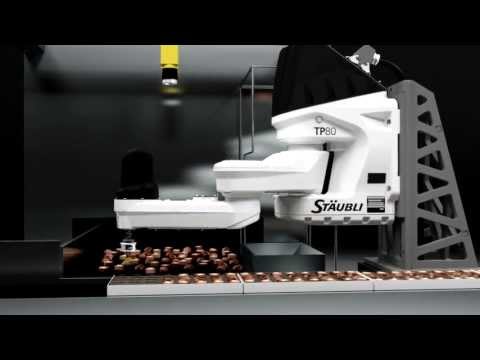}
 \caption{Examples of robots deployed in industrial settings tasked to
   arrange objects in desired configurations through pick and place:
   (left) ABB's IRB 360 FlexPicker rearranging pancakes (right)
   St\"{a}ubli's TP80 Fast Picker robot.}
 \label{fig:automation}
\end{figure}

In many industrial and logistics applications, such as those shown in
Fig. \ref{fig:automation}, a robot is tasked to rearrange multiple,
similar objects placed on a tabletop into a desired arrangement. In
these setups, the robot needs to approach the objects from above and
perform a \textit{pick-and-place action} at desired target poses. Such
operations are frequently part of product packaging and inspection
processes. Efficiency plays a critical role in these domains, as the
speed of task completion has a direct impact on financial viability.
Beyond industrial robotics, a home assistant robot may need to deal
with such problems as part of a room cleaning task. The reception of
such a robot by people will be more positive if its solutions are
efficient and does not waste time performing redundant actions. Many
subtasks affect the efficiency of the overall solution in all of these
applications, ranging from perception to the robot's speed in grasping
and transferring objects. But overall efficiency critically depends on
the underlying combinatorial aspects of the problem, which relate to
the number of pick-and-place actions that the robot performs, the
placement of the objects, as well as the sequence of objects
transferred.

This paper deals with the combinatorial aspects of such object
rearrangement tasks. The objective is to understand the underlying
structure and obtain high-quality solutions in a computationally
efficient manner. The focus is on a subset of general rearrangement
problems, which relate to the above mentioned applications. In
particular, the setup corresponds to rearranging multiple,
similar-geometry, non-stacked objects on a flat, horizontal surface
from given initial to target arrangements. The robot can approach the
objects from above, pick them up and raise them. At that point, it can
move them freely without collisions with other objects.

%

\begin{figure}[t]
  
    \resizebox{\columnwidth}{!}{
    \includegraphics[height=1.2in]{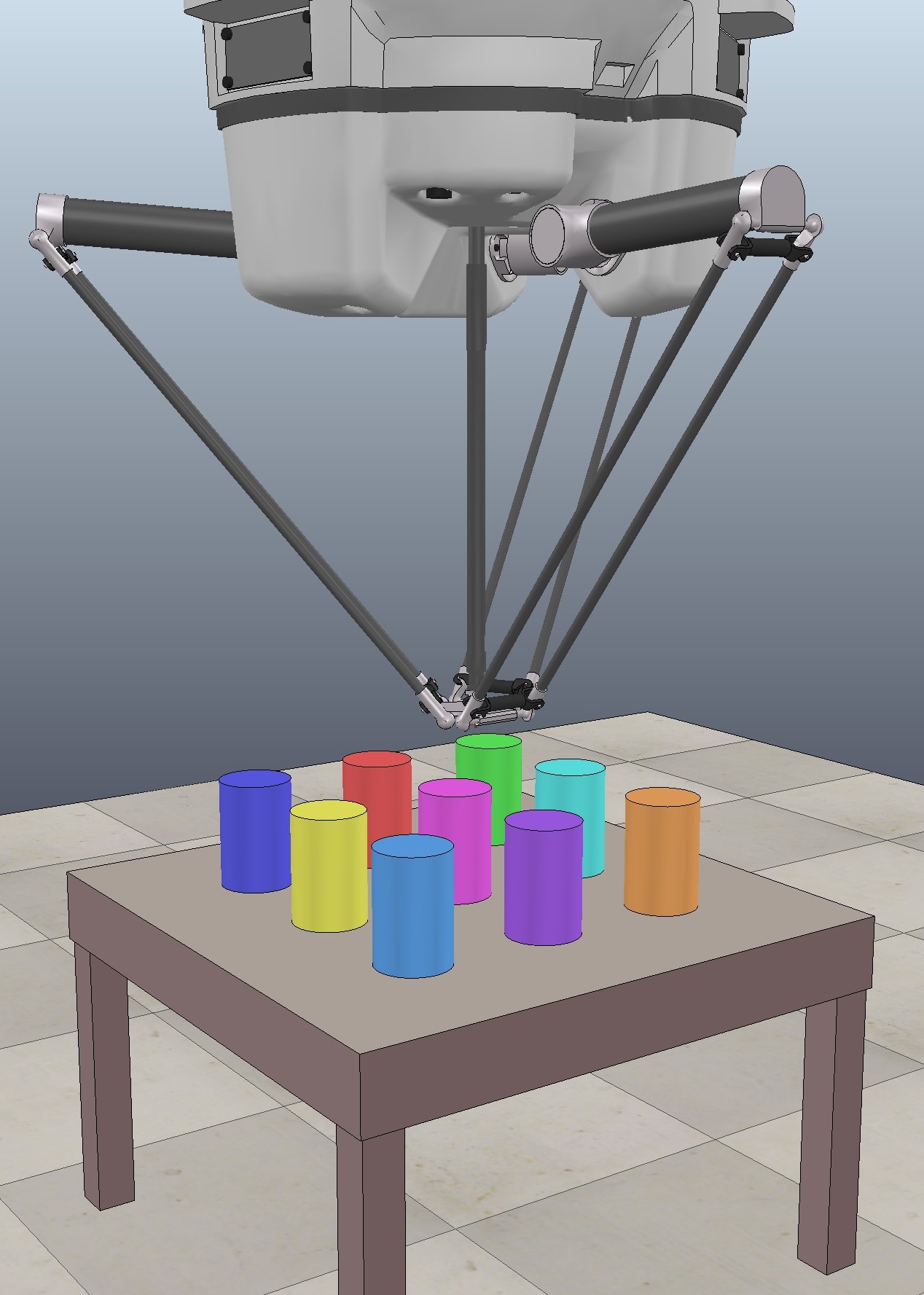}
    \includegraphics[height=1.2in]{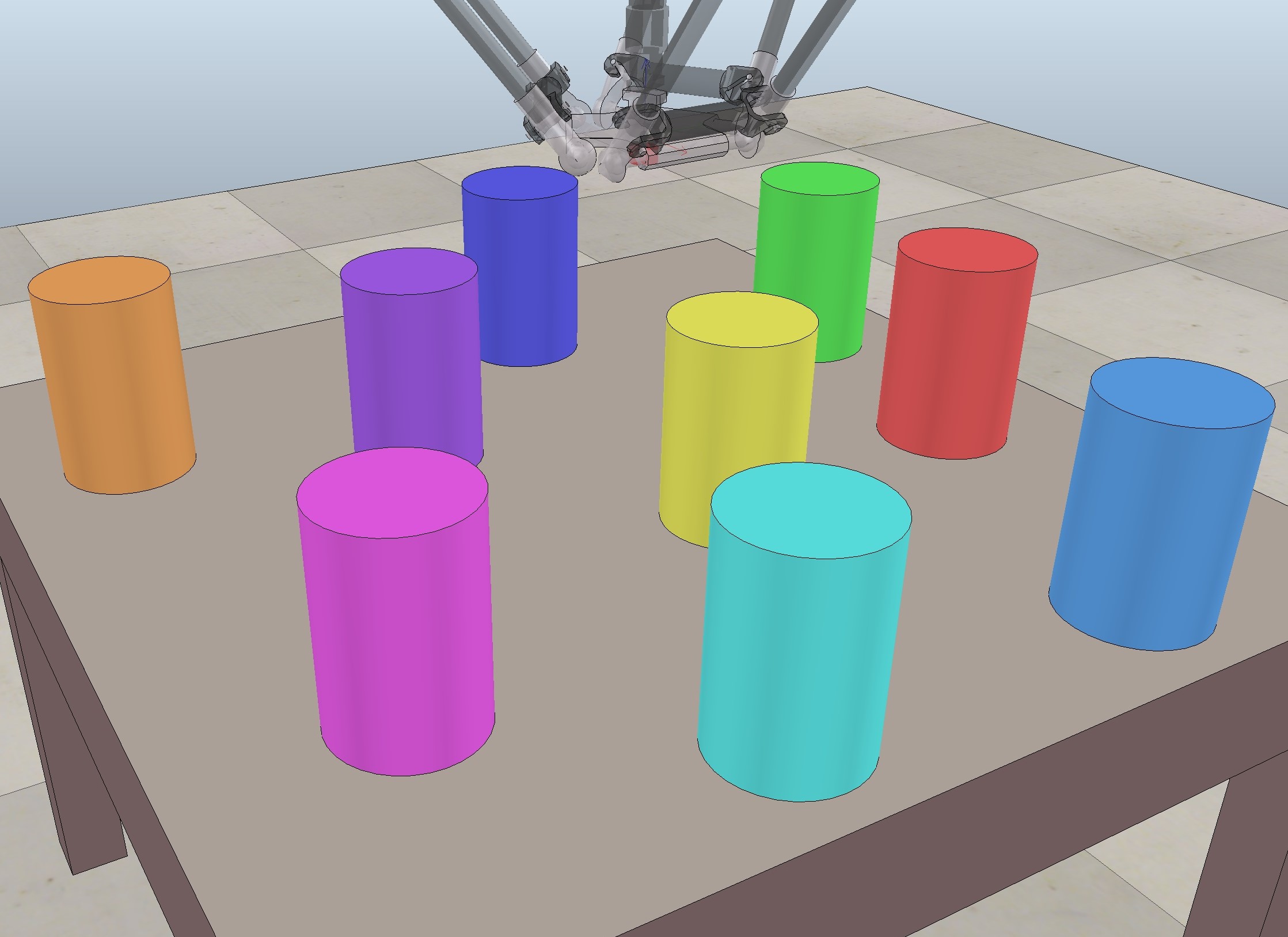}
    \includegraphics[height=1.2in]{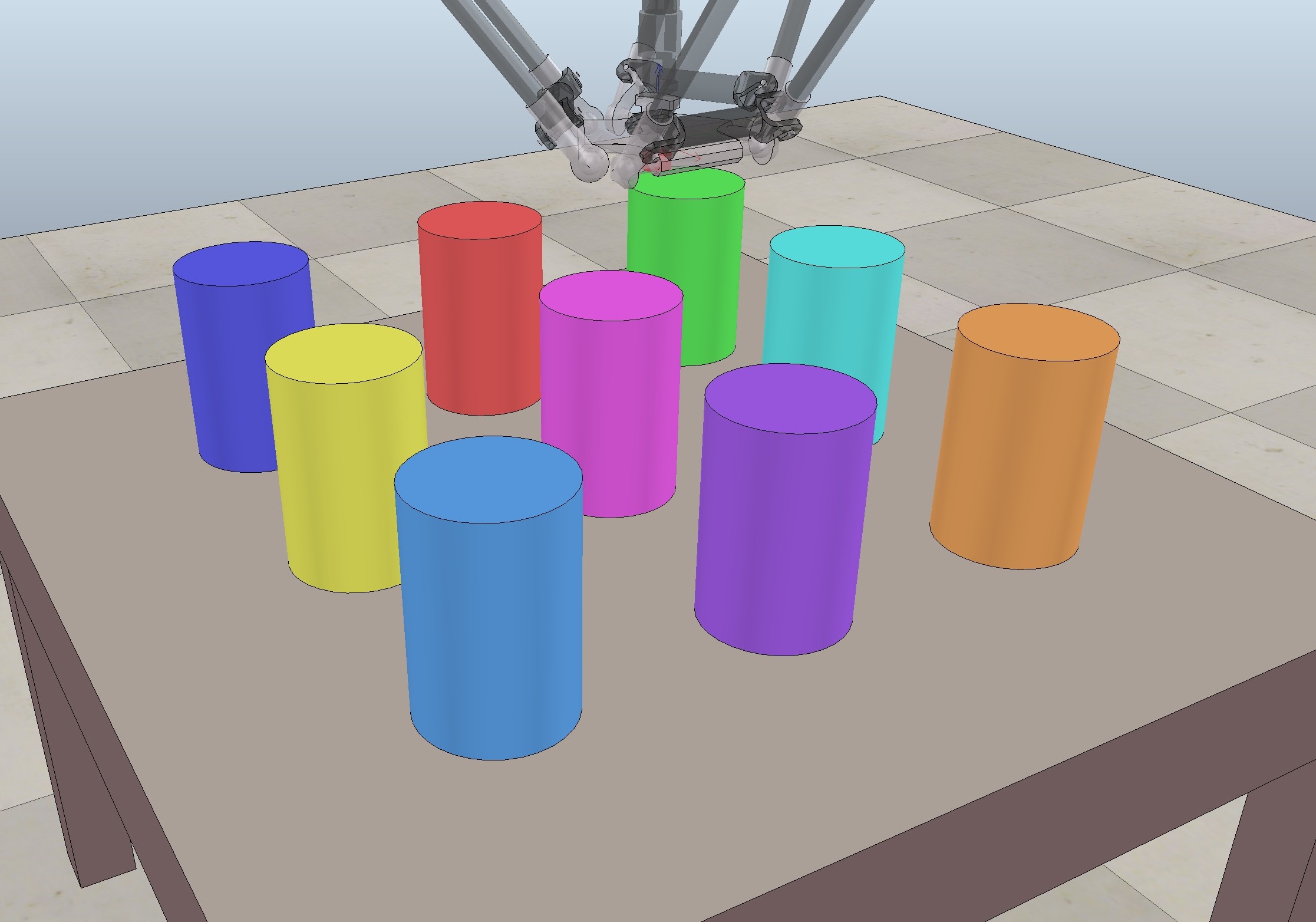}
    }
   \caption{An example of an object rearrangement challenge considered
     in this work where the initial (middle) and final (right) object
     poses are overlapping and an object needs to be placed at an
     intermediate location. Images from a V-REP~\cite{RohSinFre13}
     simulation.}
   \label{fig:example}
\end{figure}

There are two important variations of this problem. The first requires
that the target object poses do not overlap with the initial ones.  In
this scenario, the number of pick-and-place actions is equal to the
number of objects not in their goal pose. Thus, the solution quality
is dependent upon the sequence with which the objects are transferred.
A good sequence can minimize the distance that the robot's
end-effector travels.  The second variant of the problem allows for
target poses to overlap with the initial poses, as in Fig.
\ref{fig:example}. The situation sometimes necessitates the
identification of intermediate poses for some objects to complete the
task. In such cases, the quality of the solution tends to be dominated
by the number of intermediate poses needed to solve the problem, which
correlates to the number of the pick-and-place actions the robot must
carry out. The primary objective is to find a solution, which uses the
minimum number of intermediate poses and among them minimize the
distance the robot's end-effector travels.

Both variations include some assumptions that simplify these instances
relative to the general rearrangement problem.  The non-overlapping
case in particular seems to be quite easy since a random feasible
solution can be trivially acquired. Nevertheless, this paper shows
that even in this simpler setup, the optimal variant of the problem
remains computationally hard. This is achieved by reducing the
Euclidean-{\tt TSP} problem \cite{Pap77} to the cost-optimal,
non-overlapping tabletop object rearrangement problem. Even in the
unlabeled case, where objects can occupy any target pose, the problem
is still hard. For overlapping initial and final poses, the paper
employs a graphical representation from the literature
\cite{BerSno+09}, which leads to the result that finding the minimum
number of pick-and-place actions relates to a well-known problem in
the algorithmic community, the ``Feedback Vertex Set'' ({\tt FVS})
problem \cite{Kar72}. This again indicates the hardness of the
challenge.

The benefit of these two-way reductions, beyond the hardness results
themselves, is that they suggest algorithmic solutions and provide an
expectation on the practical efficiency of the methods. In particular,
Euclidean-{\tt TSP} admits a polynomial-time approximation scheme
({\tt PTAS}) and good heuristics, which implies very good practical
solutions for the non-overlapping case.  On the other hand, the {\tt
  FVS} problem is {\tt APX}-hard \cite{Kar72, DinSaf05}, which
indicates that efficient algorithms are harder for the overlapping
case.  This motivated the consideration of alternative heuristics for
solving such challenges that make sense in the context of object
rearrangement.

The algorithms proposed here, which arose by mapping the object
rearrangement variations to established, well-studied problems, have
been evaluated in terms of practical performance. For the
non-overlapping case, an alternative solver exists that was developed
for a related challenge \cite{TrePavFra13}. The {\tt TSP} solvers
achieve superior performance relative to this alternative when applied
to object rearrangement.  They achieve sub-second solution times for
hundreds of objects. Optimal solutions are shown to be significantly
better than the average, random, feasible solution.  For the
overlapping case, exact and heuristic solvers are considered.  The
paper shows that practically efficient methods achieve sub-second
solution times without a major impact in solution quality for tens of
objects.

\section{Contribution Relative to Prior Work}
\label{sec:related_work}

\emph{Multi-body planning} is a related challenge that is itself
hard. In the general, continuous case, complete approaches do not
scale even though methods exist that try to decrease the effective
DOFs \cite{AroBer+99}. For specific geometric setups, such as
unlabeled unit-discs among polygonal obstacles, optimality can be
achieved \cite{SolYu+15}, even though the unlabeled case is still hard
\cite{SolHal15}. Given the hardness of multi-robot planning, decoupled
methods, such as priority-based schemes \cite{BerOve05} or velocity
tuning \cite{LerLauSim99}, trade completeness for efficiency. Assembly
planning \cite{WilLat94, HalLatWil00, SunRemAma01} deals with similar
problems but few optimality arguments have been made.

Recent progress has been achieved for the discrete variant of the
problem, where robots occupy vertices and move along edges of a
graph. For this problem, also known as ``pebble motion on a graph''
\cite{KorMilSpi84, CalDumPac08, AulMon+99, GorHas10}, feasibility can
be answered in linear time and paths can be acquired in polynomial
time. The optimal variation is still hard but recent optimal solvers
with good practical efficiency have been developed either by extending
heuristic search to the multi-robot case \cite{WagKanCho12,
  SharSte+15}, or utilizing solvers for other hard problems, such as
network-flow \cite{YuLaV12, YuLaV16}. The current work is motivated by
this progress and aims to show that for certain useful rearrangement
setups it is possible to come up with practically efficient algorithms
through an understanding of the problem's structure.

\emph{Navigation among Movable Obstacles} ({\tt NAMO}) is a related
computationally hard problem \cite{Wil91, CheHwa91, DemORoDem00,
  NieStaOve06}, where a robot moves and pushes objects.  A
probabilistically complete solution exists for this problem
\cite{BerSti+08}.  {\tt NAMO} can be extended to manipulation among
movable obstacles ({\tt MAMO}) \cite{StiSch+07} and rearrangement
planning \cite{BenRiv98, Ota04}. Monotone instances for such problems,
where each obstacle may be moved at most once, are easier
\cite{StiSch+07}. Recent work has focused on ``non-monotone''
instances \cite{HavOzb+14, SriFan+14, GarLozKae14, KroSho+14, KroBek15a,
  KroBek16}. Rearrangement with overlaps considered in the current
paper includes ``non-monotone'' instances although other aspects of
the problem are relaxed. In all these efforts, the focus is on
feasibility and no solution quality arguments have been provided.
Asymptotic optimality has been achieved for the related ``minimum
constraint removal'' path problem \cite{Hau14, Hau13}, which, however,
does not consider negative object interactions.

The \emph{Pickup and Delivery Problem} ({\tt PDP}) \cite{BerCor+07,
  BerCorLap10} is a well-studied problem in operations research that
is similar to tabletop object rearrangement, as long as the object
geometries are ignored.  The {\tt PDP} models the pickup and delivery
of goods between different parties and can be viewed as a subclass of
vehicle routing \cite{Lap92} or dispatching \cite{ChrEil69}.  It is
frequently specified over a graph embedded in the {\tt 2D} plane,
where a subset of the vertices are pickup and delivery locations. A
{\tt PDP} in which pickup and delivery sites are not uniquely paired
is also known as the NP-hard swap problem \cite{GarJoh79, AniHas92},
for which a 2.5-optimal heuristic is known \cite{AniHas92}.  Many
exact linear programming algorithms and approximations are available
\cite{BeuOudWas04, HofLok06, GriHal+07} when pickup and delivery
locations overlap, where pickup must happen some time after delivery.
The stacker crane problem ({\tt SCP}) \cite{FreHecKim76, TrePavFra13}
is a variation of {\tt PDP} of particular relevance as it maps to the
non-overlapping case of labeled object rearrangement.  An
asymptotically optimal solution for {\tt SCP} \cite{TrePavFra13} is
used as a comparison point in the evaluation section.





This work does not deal with other aspects of rearrangement, such as
arm motion \cite{SimLau+04, BerSriKuf12, CohChiLik13, ZucRat+13} or
grasp planning \cite{CioAll09, BohMor+14}.  Non-prehensile actions,
such as pushing, are also not considered \cite{CosHer+11,
  DogSri11}. Similar combinatorial issues to the ones studied here are
also studied by integrated task and motion planners, for most of which
there are no optimality guarantees \cite{CamAlaGra09, PlaHag10,
  SriFan+14, GarLozKae14, GhaLalAla15, DanKinCha+16}.  Recent work on
asymptotically optimal task planning is at this point prohibitively
expensive for practical use \cite{VegRoy16}.

\section{Problem Statement}
\label{sec:preliminaries}
	
This section formally defines the considered challenges.
	
\subsection{Tabletop Object Rearrangement with Overhand Grasps}

Consider a workspace $\Wspace$ with static obstacles and a set of $n$
movable objects $\objects = \{o_1, \dots,o_n\}$. For $o_i \in
\objects$, $\Cspace_i$ denotes its configuration space. Then,
$\Fspace_i \subseteq \Cspace_i$ is the set of collision-free
configurations of $o_i$ with respect to the static obstacles in
$\Wspace$. An {\em arrangement} $R = \{r_1, \dots, r_n\}$ for the
objects $\objects$ specifies the configurations $r_i \in \Cspace_i$
for each object $o_i$.  A feasible arrangement is one satisfying:
\begin{enumerate}
\item $\forall\ r_i \in R, r_i \in \Fspace_i$;
\item $\forall\ r_i, r_j \in R$, if $i \neq j$, then objects $o_i$ and
  $o_j$ are not \changed{in collision} when placed at $r_i$ and $r_j$,
  respectively.
\end{enumerate}

This work focuses on bounded planar workspaces: $\mathcal W \in
\mathbb R^2$. The setting is frequently referred to as the {\em
  tabletop} case, in which the vertical projections of the objects on
the tabletop do not intersect. This work assumes that the manipulator
is able to employ {\em overhand} grasps, where an object can be
transferred after being lifted above all other objects.  In
particular, a pick-and-place operation of the manipulator involves
four steps:
\begin{enumerate}[\hspace*{1em}a.]
    \item bringing the end-effector above the object,
    \item \changed{grasping} and \changed{lifting} the object,
    \item transfer \changed{of} the
        grasped object horizontally to its target (horizontal) location, and
    \item \changed{a downward motion prior} to releasing the object. 
\end{enumerate}
This sequence is defined as a {\em manipulation action}.

The manipulator is initially at a rest position $s_M$ prior to
executing any pick-and-place actions and transitions to a rest
position $g_M$ at the conclusion of the rearrangement task. A {\em
  rest position} is a safe arm configuration, where there is no
collision with objects.

The illustrations that appear throughout the paper assume objects with
identical geometry. Nevertheless, the results derived in this paper
are not dependent on this assumption, i.e., objects need only be
cylindrical in diff. geometry terms.\footnote{From differential
  geometry, a cylinder is defined as any ruled surface spanned by a
  one-parameter family of parallel lines.}

Given the above assumptions, the problem studied in the paper can be
summarized as:
\begin{problem}
\textbf{Tabletop Object Rearrangement with Overhand grasps (\toro).}
Given feasible start and goal arrangements $R_S, R_G$ for objects
$\objects = \{o_1, \dots,o_n\}$ on a tabletop, determine a sequence of
collision-free pick-and-place actions with overhand grasps $\actions =
(a^1,a^2, \dots)$ that transfer $\objects$ from $R_S$ to $R_G$.
\end{problem}

A rearrangement problem is said to be {\em labeled} if objects are
unique and not interchangeable. Otherwise, the problem is {\em
  unlabeled}. If for two arbitrary arrangements $s \in R_S$ and $g \in
R_G$, the objects placed in $s$ and $g$ are not in collision, then the
problem is said to have {\em no overlaps}.  Otherwise, the problem is
said to have {\em overlaps}.


This paper primarily focuses on the labeled \toro case and identifies
an important subcase:
\begin{itemize}
  \item \textit{\toro with NO overlaps (\rno)}
\end{itemize}

\remark The partition of Problem 1 into the general \rwo case and the
subcase of \rno is not arbitrary. \rwo is structurally richer and
harder from a computational perspective.  Both version of the problem
can be extended to the unlabeled and partially labeled variants.  This
paper does not treat the labeled and unlabeled variants as separate
cases but will briefly discuss differences that arise due to
formulation when appropriate.


\subsection{Optimization Criteria}
Recall that a manipulation action $a^i$ has four components: an initial
move, a grasp, a transport phase, and a release. Since grasping is
frequently the source of difficulty in object manipulation tasks, it
is assumed in the paper that grasps and releases induce the most cost
in manipulation actions. The other source of cost can be attributed to
the length of the manipulator's path. This part of the cost is
captured through the Euclidean distance traveled by the end effector
between grasps and releases. For a manipulation action $a^i$, the
incurred cost is
\begin{equation}\label{equ:action}
c_{a^i} = c_md^i_e + c_g + c_md^i_l + c_r,
\end{equation}
\changed{where} $c_m,~c_g,~c_r$ are costs associated with moving the
manipulator, a single grasp, and a single release,
respectively. $d_e^i$ and $d_l^i$ are the straight line distances
traveled by the end effector in the first (object-free) and third
(carrying an object) stages of a manipulation action, respectively.
		
The total cost associated with solving a \toro instance is then
captured by
\vspace{0.01in}
\begin{equation}\label{equ:totalcost}
c_{T} = \sum_{i = 1}^{|\actions|} c_{a^i} = |\actions| (c_g + c_r) + c_m\Big(\sum_{i = 1}^{|\actions|} (d^i_e + d^i_l) + d_f\Big),
\end{equation}
\vspace{0.01in}
\noindent where $d_f$ is the distance between the location of the last
release of the end effector and its rest position $g_M$. Of the two
additive terms in~\eqref{equ:totalcost}, note that the first term
dominates the second.  Because the absolute value of $c_g, c_r$, and
$c_m$ are different for different systems, the assignment of their
absolute values is left to practitioners. The focus of this paper is
the analysis and minimization of the two additive terms
in~\eqref{equ:totalcost}.

\subsection{Object Buffer Locations}
The resolution of \rwo (Section~\ref{sec:rwo}) may require the
temporary placement of some object(s) at intermediate locations
outside those in $R_S \cup R_G$. When this occurs, external buffer
locations may be used as temporary locations for object
placement. More formally, there exists a set of configurations $B =
\{b_1, b_2, \dots\}$, called \textsl{buffers}, which are available to
the manipulator and do not overlap with object placements in $R_S$ or
$R_G$. 


\remark This work, which focuses on the combinatorial aspects of
multi-object manipulation and rearrangement, utilizes exclusively
buffers that are not on the tabletop. It is understood that the number
of external buffers \textit{may} be reduced by attempting to first
search for potential buffers within the tabletop. Nevertheless, there
are scenarios where the use of external buffers may be necessary.


\section{TORO with No Overlaps (\rno)}\label{sec:rno}
When there is no overlap between any pair of start and goal configurations, 
an object can be transferred directly from its start configuration to its
goal configuration. A direct implication is that an optimal sequence of 
manipulation actions contains exactly $|\actions| = |\objects| = n$ grasps and 
the same number of releases. Note that a minimum of $n$ grasps and 
releases are necessary. This also implies that no buffer is required 
since using buffers will incur additional grasp and release costs. 
Therefore, for \rno,~\eqref{equ:totalcost} becomes 
\begin{equation}\label{equ:tcno}
c_{T} = n(c_g + c_r) + c_m\Big(\sum_{i = 1}^{n} (d^i_e + d^i_l) + d_f\Big),
\end{equation}
i.e., only the distance traveled by the end effector affects the cost. 
The problem instance that minimizes~\eqref{equ:tcno} is referred to as Cost-optimal \rno.
The following theorem provides a hardness result for Cost-optimal \rno.

\begin{theorem}\label{t:lno-np-hard}Cost-optimal \rno\ is NP-hard.  
\end{theorem} 
\noindent\begin{IEEEproof}
Reduction from Euclidean-\tsp\ \cite{Pap77}. Let $p_0,p_1,\ldots,
p_n$ be an arbitrary set of $n+1$ points in 2D. The set of points 
induces an Euclidean-\tsp. Let $d_{ij}$ denote the Euclidean distance 
between $p_i$ and $p_j$ for $0 \le i, j \le n$. In the formulation 
given in \cite{Pap77}, it is assumed that $d_{ij}$ are integers, which 
is equivalent to assuming the distances are rational numbers. To reduce 
the stated \tsp problem to a cost-optimal \rno\ problem, pick some 
positive $\varepsilon \ll 1/(4n)$. Let $p_0$ be the rest position of 
the manipulator in an object rearrangement problem. For each $p_i$, 
$1 \le i \le n$, split $p_i$ into a pair of start and goal configurations 
$(s_i, g_i)$ such that {\it (i)}$p_i = \frac{s_i + g_i}{2}$, {\it (ii)} 
$s_{i2} = g_{i2}$, and {\it (iii)} $s_{i1} + \varepsilon = g_{i1}$. An 
illustration of the reduction is provided in Fig.~\ref{fig:rno-reduction}. 
The reduced \rno instance is fully defined by $p_0$, $R_S = \{s_1, \ldots,
s_n\}$ and $R_G = \{g_1, \ldots, g_n\}$. A cost-optimal (as defined 
by~\eqref{equ:tcno}) solution to this \rno\ problem induces a (closed) 
path starting from $p_0$, going through each $s_i$ and $g_i$ exactly once, 
and ending at $p_0$. Moreover, each $g_i$ is visited immediately
after the corresponding $s_i$ is visited. Based on this path, the manipulator
moves to a start location to pick up an object, drop the object at the
corresponding goal configuration, and then move to the next object until all
objects are rearranged. Denote the loop path as $P$ and let its total length be 
$D$.

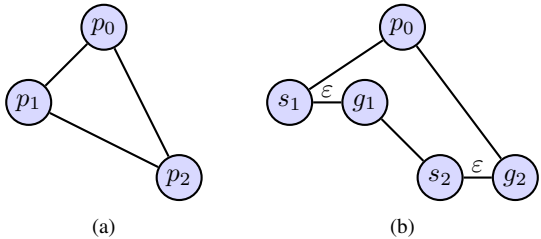
\begin{figure}[htp]
	\begin{center}
	\begin{tabular}{ccc}
	\begin{tikzpicture}[scale=1]
		\foreach \nodeName/\nodeLocation in {p_0/{(0, 2)}, p_1/{(-1, 1)}, p_2/{(1, 0)}}{
			\node (\nodeName) at \nodeLocation {$\nodeName$};
		}
		\foreach \edgeFrom/\edgeTo in {p_0/p_1, p_1/p_2, p_2/p_0}{
			\draw [-] (\edgeFrom) to (\edgeTo);
		}
	\end{tikzpicture}
	&&
	\begin{tikzpicture}[scale=1]
		\foreach \nodeName/\nodeLocation in {p_0/{(0, 2)}, s_1/{(-1.5, 1)}, 
		g_1/{(-0.5, 1)}, s_2/{(0.5, 0)}, g_2/{(1.5, 0)}}{
			\node (\nodeName) at \nodeLocation {$\nodeName$};
		}
		\foreach \edgeFrom/\edgeTo in {p_0/s_1, s_1/g_1, g_1/s_2, s_2/g_2, g_2/p_0}{
			\draw [-] (\edgeFrom) to (\edgeTo);
		}
		\node[draw=none, fill=none] at (1, 0.15) {$\varepsilon$};
		\node[draw=none, fill=none] at (-1, 1.15) {$\varepsilon$};
	\end{tikzpicture}\\
	{\footnotesize (a)} && {\footnotesize (b)}\\
	\end{tabular}
	\end{center}
	\caption{\label{fig:rno-reduction} \changed{Reduction from Euclidean-\tsp{} to cost-optimal \rno{}}}  
\end{figure}

Assume that the Euclidean-\tsp has an optimal solution path $P_{opt}$ 
with a total distance of $D_{opt}$ (an integer). Then $P$ from solving the cost-optimal 
\rno\ yields such an optimal path for the \tsp. To show this, from $P$, 
simply contract the edges $s_ig_i$ for all $1 \le i \le n$. This clearly 
yields a solution to the Euclidean-\tsp; let the resulting path be $P'$ 
with total length $D'$. As edges are contracted along $P$, by the triangle 
inequality, $D' \le D$. It remains to show that $D' = D_{opt}$. Suppose this 
is not the case, then $D' \ge D_{opt} + 1$. However, if this is the case, 
a solution to the \rno can be constructed by splitting $p_i$ into $s_i$ and $g_i$ 
along $P_{opt}$. It is straightforward to establish that the total distance
of this \rno path is bounded by $D_{opt} + n\varepsilon < D_{opt} 
+ n*1/(4n) = D_{opt} + 1/4 < D_{opt} + 1 \le D' \le D$. Since this is a 
contradiction, $D' = D_{opt}$. 
\end{IEEEproof}

\remark Note that an NP-hardness proof of a similar problem can be
found in \cite{FreGua93}, as is mentioned in
\cite{TrePavFra13}. Nevertheless, the problem is stated for a tree and
is non-Euclidean. Furthermore, it is straightforward to show that the
decision version of the cost-optimal \rno\ problem is NP-complete;
this non-essential detail is omitted.

\remark Interestingly, \rno\ also reduces to \tsp with very little
effort.  Because highly efficient \tsp solvers are available, the
reduction route provides an effective approach for solving
\rno{}. This is not always a feature of NP-hardness reductions. The
straightforward algorithm for the computation is outlined in
Alg.~\ref{algo:lno}. The inputs to the algorithm are the rest
positions of the manipulator and the start and goal configurations of
objects. The output is the solution for \rno, represented as a
sequence of manipulation actions $\actions$, which has completeness
and optimality guarantees.

\def\rnos{{\sc ToroNoTSP}\xspace}
\begin{algorithm}
\begin{small}
	\DontPrintSemicolon
	\KwIn{Configurations $s_M$, $g_M$, Arrangements $R_S$, $R_G$.}
	\KwOut{A sequence of manipulation actions $\actions$.}	
	$G_{NO} \leftarrow ${\sc ConstructTSPGraph}$(R_S, R_G, s_M, s_G)$\;\label{algo:lno_model}
	$S_{raw}  \gets$ {\sc SolveTSP}$(G_{NO})$\;\label{algo:lno_solve}
	$\actions \gets$ {\sc RetrieveActions}$(S_{raw})$\;\label{algo:rno_action}
	\Return{$\actions$}\;
	\caption{{\sc ToroNoTSP}}
	\label{algo:lno}
\end{small}
\end{algorithm}
	 
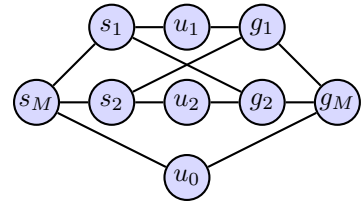
\begin{figure}[htp]
		\centering
		\begin{tikzpicture}[scale=1]
			\foreach \nodeName/\nodeLocation in {s_M/{(0, 0)}, s_1/{(1, 1)}, 
			s_2/{(1, 0)}, u_1/{(2, 1)}, u_2/{(2, 0)}, g_1/{(3, 1)}, 
			g_2/{(3, 0)}, g_M/{(4, 0)}, u_0/{(2, -1)}}{
				\node (\nodeName) at \nodeLocation {$\nodeName$};
			}
			\foreach \edgeFrom/\edgeTo in {s_M/s_1, s_1/u_1, u_1/g_1, g_1/g_M, 
			s_M/s_2, s_2/u_2, u_2/g_2, g_2/g_M, s_M/u_0, u_0/g_M, s_1/g_2, s_2/g_1}{
				\draw [-] (\edgeFrom) to (\edgeTo);
			}
		\end{tikzpicture}
		\caption{An example of $G_{NO}$ for 2 objects. The nodes 
		$s_M$ and $g_M$ denote the initial and final rest position of the 
		manipulator end effector.}
		\label{fig:rno-tsp}
	\end{figure}

At Line \ref{algo:lno_model} of Alg.~\ref{algo:lno}, a graph
$G_{NO}(V_{NO}, E_{NO})$ is generated as the input to the \tsp
problem.  The graph is constructed from the \rno\ instance as
follows. A vertex is created for each element of $R_S$ and
$R_G$. Then, a complete bipartite graph is created between these two
sets of vertices. A set of vertices $U = \{u_1, \ldots, u_{|R_S|}\}$
is then inserted into edges $s_ig_i$ for $1 \le i \le
|R_S|$. Afterward, $s_M$ (resp., $g_M$) is added as a vertex and is
connected to $s_i$ (resp., $g_i$) for $1 \le i \le |R_S|$. Finally, a
vertex $u_0$ is added and connected to both $s_M$ and $g_M$. See
Fig.~\ref{fig:rno-tsp} for the straightforward example for $|R_S| =
2$.
	
Let $w(a, b)$ denote the weight of an edge $(a, b) \in E_{NO}$. For 
all $1 \leq i, j \leq n, i \neq j$ ($\text{dist}(x, y)$ denotes the	
Euclidean distance between $x$ and $y$ in 2D):
	\begin{align*}
		& w(s_M, u_0) = w(g_M, u_0) = 0, 
		w(s_M, s_i) = \text{dist}(s_M, s_i), \\
		& w(g_M, g_i) = \text{dist}(g_M, g_i), 
		w(s_i, u_i) = w(u_i, g_i) = 0, \\
		& w(s_i, g_j) = \text{dist}(s_i, g_j).	
	\end{align*}
With the construction, a \tsp tour through $G_{NO}$ must use
$s_Mu_0g_M$ and all $s_iu_ig_i$ for all $1 \le i \le |R_S|$. To form a
complete tour, exactly $(|R_S| - 1)$ edges of the form $g_is_j$, where
$i \ne j$ must be used. At Line \ref{algo:lno_solve}, the \tsp is
solved (Concorde \tsp solver \cite{AppBix+07} is used). This gives a
minimum weight solution $S_{raw}$, which is a cycle containing all $v
\in V_{NO}$. The manipulation actions can then be retrieved (Line
\ref{algo:rno_action}).

An alternative solution to \rno\ could employ the asymptotically optimal, \textsl{SPLICE} algorithm, 
introduced in \cite{TrePavFra13}.
	

The scenario where objects are 
unlabeled is a special case of \rno which has significance in real-world applications (e.g., the 
pancake stacking application). This case is denoted as \uno\ (unlabeled, 
no overlap). Adapting the NP-hardness proof for the \rno\ problem 
shows that cost-optimal \uno\ is also NP-hard. Similar to the \rno\ 
case, the optimal solution only hinges on the distance traveled by the 
manipulator because no buffer is required and exactly $n$ grasps and 
releases are needed. 
\begin{theorem}\label{t:uno-np-hard}Cost-optimal \uno\ is NP-hard.  
\end{theorem} 
\begin{IEEEproof}
  See Appendix A of \cite{HanStiKroBekYu17RSSSup}.
\end{IEEEproof}

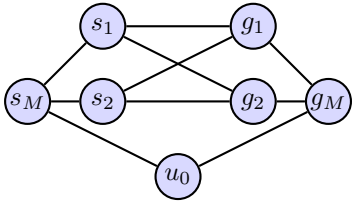
\begin{figure}[htp]
		\centering
		\begin{tikzpicture}[scale=1]
			\foreach \nodeName/\nodeLocation in {s_M/{(0, 0)}, s_1/{(1, 1)}, 
			s_2/{(1, 0)}, g_1/{(3, 1)}, g_2/{(3, 0)}, g_M/{(4, 0)}, u_0/{(2, -1)}}{
				\node (\nodeName) at \nodeLocation {$\nodeName$};
			}
			\foreach \edgeFrom/\edgeTo in {s_M/s_1, g_1/g_M, s_M/s_2, g_2/g_M, 
			s_M/u_0, u_0/g_M, s_1/g_2, s_2/g_1, s_1/g_1, s_2/g_2}{
				\draw [-] (\edgeFrom) to (\edgeTo);
			}
		\end{tikzpicture}
		\caption{An example of $G_{UNO}$ for 2 objects.}
		\label{fig:uno-tsp}
	\end{figure}

When solving a \uno instance, Alg. \ref{algo:lno} may be used with a 
few small changes. First, a different underlying graph must be constructed. 
Denote the new graph as $G_{UNO}(V_{UNO}, E_{UNO})$, where 
$V_{UNO} = R_S \cup R_G \cup \{s_M, u_0, g_M\}$. For all $1 \leq i,j \leq n$:
	\begin{align*}
		& w(s_M, u_0) = w(g_M, u_0) = 0,  w(s_M, s_i) = \text{dist}(s_M, s_i), \\
		& w(g_M, g_i) = \text{dist}(g_M, g_i), w(s_i, g_j) = \text{dist}(s_i, g_j).
	\end{align*}
All other edges are given infinite weight. An example of the updated 
structure of $G_{UNO}$ for two objects is illustrated in Fig.~\ref{fig:uno-tsp}. 
	
\section{TORO With Overlap (\rwo)}\label{sec:rwo}
Unlike \rno, \rwo\ has a more sophisticated structure and may require
buffers to solve. In this section, a {\em dependency graph}
\cite{BerSno+09} is used to model the structure of \rwo, which leads
to a classical NP-hard problem known as the {\em feedback vertex set}
problem \cite{Kar72}. The connection then leads to a complete
algorithm for optimally solving \rwo.

\subsection{The Dependency Graph and NP-Hardness of \rwo}

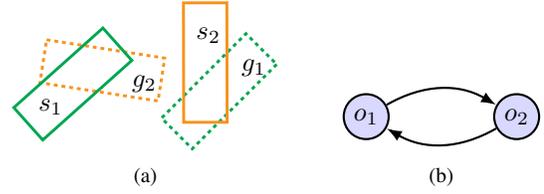
\begin{figure}
	\begin{center}
	\begin{tabular}{ccc}
  	\begin{overpic}[width=1.4in]{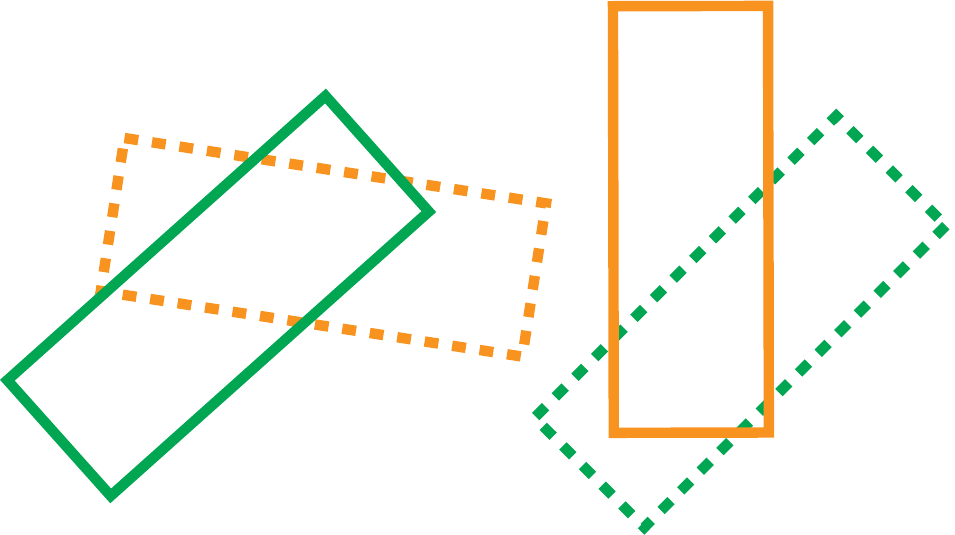}
	\put(10,15){{$s_1$}}
	\put(45,24){{$g_2$}}
	\put(69,42){{$s_2$}}
	\put(86,30){{$g_1$}}
	\end{overpic} &&
	
	\begin{tikzpicture}[scale=1]
		\node (o1) at (0, 0) {$o_1$};
		\node (o2) at (2, 0) {$o_2$};
		\draw [->] (o1) to [out = 30, in = 150] (o2);
		\draw [->] (o2) to [out = 210, in = -30] (o1);
	\end{tikzpicture}\\
			{\footnotesize (a)} && {\footnotesize (b)}\\
			\end{tabular}
			\end{center}
		\caption{Illustration of the dependency graph. (a) Two objects are to be moved 
		from $s_i$ to $g_i$, $i = 1, 2$. Due to the overlap between $s_1$ and $g_2$ as 
		well as the overlap between $s_2$ and $g_1$, one of the objects must be temporarily 
		moved aside. (b) The dependency graph capturing the scenario in (a).}
		\label{fig:dependencygraph}
	\end{figure}

Consider a {\em dependency digraph} $G_{dep}(V_{dep}, A_{dep})$, where 
$V_{dep} = \objects$, and $(o_i, o_j) \in A_{dep}$ \textit{iff} $g_i$ 
and $s_j$ overlap. Therefore, $o_j$ must be moved away from $s_j$ before 
moving $o_i$ to $g_i$. An example involving two objects is provided in 
Fig.~\ref{fig:dependencygraph}. The definition of dependency graph 
implies the following two observations. 
	
\begin{observation} If the out-degree of $o_i \in V_{dep}$ is 0, then $o_i$ 
can move to $g_i$ without collision. 
\end{observation}


\begin{observation}\label{l:lwo-dep-acyclic}
If $G_{dep}$ is not acyclic, solving \rwo\ requires at least $n + 1$ 
grasps.
\end{observation}

	
The dependency graph has obvious similarities to the well known {\em
  feedback vertex set} (\fvs) problem \cite{Kar72}. A directed \fvs
problem is defined as follows. Given a strongly connected directed
graph $G = (V, A)$, an FVS is a set of vertices whose removal leaves
$G$ acyclic. Minimizing the cardinality of this set is NP-hard, even
when the maximum in degree or out degree is no more than two
\cite{GarJoh79}.  As it turns out, the set of removed vertices in an
\fvs problem mirrors the set of objects that must be moved to
temporary locations (i.e., buffers) for resolving the dependencies
between the objects, which corresponds to the additional grasps (and
releases) that must be performed in addition to the $n$ required
grasps for rearranging $n$ objects. The observation establishes that
cost-optimal \rwo\ is also computationally intractable. The following
lemma shows this point.
\begin{lemma}\label{l:lwo-min-grasp}
Let the dependency graph of a \rwo\ problem be a single strongly connected
graph. Then the minimum number of additional grasps required for solving the
\rwo\ problem equals the cardinality of the minimum FVS of the dependency 
graph. 
\end{lemma}
\begin{IEEEproof}
Given the dependency graph, let the additional grasps and releases be $n_x$ 
and the minimum FVS have a cardinality of $n_{fvs}$,
it remains to show that $n_x = n_{fvs}$. First, if fewer than $n_{fvs}$ objects, 
corresponding to vertices of the dependency graph, are removed, then there
remains a directed cycle. By Observation~\ref{l:lwo-dep-acyclic}, this part 
of the problem cannot be solved. This establishes that $n_x \ge n_{fvs}$. 
On the other hand, once all objects corresponding to vertices in a minimum 
FVS are moved to buffer locations, the dependency graph 
becomes acyclic. This allows the remaining objects to be rearranged. This operation
can be carried out iteratively with objects whose corresponding vertices
have no incoming edges. On a directed acyclic graph (\DAG), there is always such a vertex. Moreover,
as such a vertex is removed from a \DAG, the remaining graph must still 
be a \DAG and therefore must have either no vertex (a trivial \DAG) or a 
vertex without incoming edges. 
\end{IEEEproof}

For dependency graphs with multiple strongly connected components, the
required number of additional grasps and releases is simply the sum of
the required number of such actions for the individual strongly
connected components. 


For a fixed \rwo\ problem, let $n_{fvs}$ be the cardinality of the largest
(minimal) FVS computed over all strongly connected components of its dependency
graph. Then it is easy to see that the maximum number of required buffers is no
more than $n_{fvs}$. The NP-hardness of cost-optimal \rwo\ is established using
the reduction from \fvs problems to \rwo. This is more involved than reducing
\rwo\ to \fvs because the constructed \rwo\ must correspond to an actual
\toro problem.

\begin{theorem}\label{t:lwo-min-buffer}
Cost-optimal \rwo is NP-hard. 
\end{theorem}
\begin{IEEEproof}
The \fvs problem on directed graphs is reduced to cost-optimal \rwo. An
\fvs problem is fully defined by specifying an arbitrary strongly
connected directed graph $G = (V, A)$ where each vertex has no more
than two incoming and two outgoing edges. A typical vertex
neighborhood can be represented as illustrated in
Fig.~\ref{fig:dep-temp}(a). Such a neighborhood is converted to a
dependency graph neighborhood of object rearrangement as follows. Each
of the original vertex $v_i \in V$ becomes an object $o_i$ which has
some $(s_i, g_i)$ pair as its start and goal configurations. For each
directed arc $v_iv_j$, split it into two arcs and add an
additional object $o_{ij}$. That is, create new arcs $o_io_{ij}$
and $o_{ij}o_j$ for each original arc $v_{ij}$ (see
Fig.~\ref{fig:dep-temp}(b)). This yields a dependency graph that is
again strongly connected. Two claims will be proven:
\begin{enumerate}
\item The constructed dependency graph corresponds to an object
  rearrangement problem, and
\item The minimum number of objects that must be moved away
  temporarily to solve the problem is the same as the size of the
  minimum FVS.
\end{enumerate}
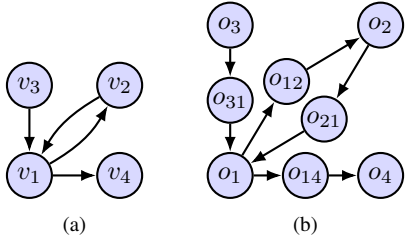
\begin{figure}[htp]
	\begin{center}
	\begin{tabular}{ccc}
	\begin{tikzpicture}[scale=1.2]
		\foreach \nodeName/\nodeLocation in {v_1/{(0, 0)}, v_2/{(1, 1)}, v_3/{(0, 1)}, v_4/{(1, 0)}}{
			\node (\nodeName) at \nodeLocation {$\nodeName$};
		}
		\foreach \edgeFrom/\edgeTo in {v_1/v_4, v_3/v_1}{
			\draw [->] (\edgeFrom) to (\edgeTo);
		}
		\draw [->] (v_1) to [out = 30, in = -120] (v_2);
		\draw [->] (v_2) to [out = -150, in = 60] (v_1);
	\end{tikzpicture}
	&&
	\begin{tikzpicture}[scale=1]
		\foreach \nodeName/\nodeLocation in {o_1/{(0, 0)}, o_2/{(2, 2)}, o_3/{(0, 2)}, 
		o_4/{(2, 0)}, o_{31}/{(0, 1)}, o_{14}/{(1, 0)}, o_{12}/{(0.75, 1.25)}, o_{21}/{(1.25, 0.75)}}{
			\node (\nodeName) at \nodeLocation {$\nodeName$};
		}
		\foreach \edgeFrom/\edgeTo in {o_3/o_{31}, o_{31}/o_1, o_1/o_{12}, o_{12}/o_2,
		 o_2/o_{21}, o_{21}/o_1, o_1/o_{14}, o_{14}/o_4}{
			\draw [->] (\edgeFrom) to (\edgeTo);
		}
	\end{tikzpicture}\\
	{\footnotesize (a)} && {\footnotesize (b)}\\
	\end{tabular}
	\end{center}
	\caption{\label{fig:dep-temp} Converting a neighborhood of a graph 
	for an \fvs problem to parts of a dependency graph for a 
	\rwo problem.}  
\end{figure}
To prove the first claim, assume without loss of generality that the
objects have the same footprints on the tabletop.  Furthermore, only
the neighborhood of $o_1$ needs to be inspected because it is isolated
by the newly added objects. Recall that an incoming edge to $o_1$
means that the start configuration $o_1$ blocks the goals of some
other objects, in this case $o_{21}$ and $o_{31}$. This can be readily
realized by putting the goal configurations of $o_{21}$ and $o_{31}$
close to each other and have them overlap with the start configuration
of $o_1$. Note that the goal configurations of $o_{21}$ and $o_{31}$
have no other interactions. Therefore, such an arrangement is always
achievable for even simple (e.g., circular or square) footprints.
Similarly, for the outgoing edges from $o_1$, which mean other objects
block $o_1$'s goal, in this case $o_{12}$ and $o_{14}$, place the
start configurations of $o_{12}$ and $o_{14}$ close to each other and
make both overlap with the goal configuration of $o_1$. Again, the
start configurations of $o_{12}$ and $o_{14}$ have no other
interactions.

The second claim directly follows Lemma~\ref{l:lwo-min-grasp}. Now,
given an optimal solution to the reduced \rwo\ problem, it remains to
show that the solution can be converted to a solution to the original
\fvs problem. The solution to the \rwo\ problem provides a set of
objects that are moved to temporary locations. This yields a minimum
FVS on the dependency graph but not the original graph $G$. Note that
if a newly created object (e.g., $o_{ij}$) is moved to a temporary
place, either object $o_i$ or $o_j$ can be moved since this will
achieve no less in disconnecting the dependency graph. Doing this
across the dependency graph yields a minimum FVS for $G$.
\end{IEEEproof}
\remark It is possible to prove that \rwo\ is NP-hard using a similar
proof to the \rno\ case. To make the proof for
Theorem~\ref{t:lno-np-hard} work here, each $p_i$ can be split into an
overlapping pair of start and goal. Such a proof, however, would bury
the structure of \rwo, which is a more difficult problem. Unlike the
Euclidean-\tsp problem, which admits $(1+\varepsilon)$-approximations
and good heuristics, \fvs problems are {\tt APX}-hard
\cite{Kar72,DinSaf05}.

\subsection{Algorithmic Solutions for \rwo}
\def\lwoalgs{{\sc ToroFVSSingle}\xspace}
\subsubsection*{Feasible algorithm}
Once the link between a \rwo\ buffer requirement and \fvs is
established, an algorithm for solving \rwo\ becomes possible.  To do
this, an FVS set is found. Then the optimal rearrangement distance is
computed for this FVS set. The procedure for doing this is outlined in
\lwoalgs (Alg. \ref{algo:lwo}).  At Line \ref{algo:lwo-gdep}, the
dependency graph $G_{dep}$ is constructed.  At Line
\ref{algo:lwo-fvs1}-\ref{algo:lwo-fvs2}, an FVS is obtained for each
strongly connected component (SCC) in $G_{dep}$. Note that if these
FVSs are optimal, then the step yields the minimum number of required
grasps (and releases) as: $\min |\actions| = n + |B|.$

The residual work is to find the solution with $n + |B|$ grasps and the 
shortest travel distance (Line~\ref{algo:lwo-model}). The manipulation
actions
are then retrieved and returned.

\begin{algorithm}
\begin{small}
	\DontPrintSemicolon
	\KwIn{Configurations $s_M$, $g_M$, Arrangements $R_S$, $R_G$}
	\KwOut{A set of manipulation actions $\actions$}	
	$G_{dep} \leftarrow ${\sc ConstructDepGraph}$(R_S, R_G)$\;\label{algo:lwo-gdep}
	$B \gets \emptyset$\;
	\For{each SCC \textbf{in} $G_{dep}$}{\label{algo:lwo-fvs1}
		$B \gets B\;\cup\;${\sc SolveFVS}$(\text{SCC})$\;\label{algo:lwo-fvs2}
	}
	$S_{raw} \gets ${\sc MinDist}$(s_M, g_M, R_S, R_G, G_{dep},B)$\; \label{algo:lwo-model}
  $\actions \gets ${\sc RetrieveActions}$(S_{raw})$\;
	\Return{$\actions$}\;
	\caption{{\sc ToroFVSSingle}}
	\label{algo:lwo}
\end{small}
\end{algorithm}


The paper explores two exact and three approximate methods as implementations
of {\sc SolveFVS}() (Line~\ref{algo:lwo-fvs2} of Alg.~\ref{algo:lwo}).  The two
exact methods are both based on integer linear programming (ILP) models,
similar to those introduced in \cite{BahSchNeu15}. They differ \changed{in} how
cycle constraints are encoded: one uses \changed{a} polynomial number of
constraints and the other simply \changed{enumerates} all possible cycles.
Denote these two exact methods as {\bf ILP-Constraint} and {\bf ILP-Enumerate},
respectively. The details of these two exact methods are explained in 
Appendix B of \cite{HanStiKroBekYu17RSSSup}. \changed{With regards to approximate solutions, several heuristic solutions are
presented}:
\begin{enumerate}
\item {\bf Maximum Simple Cycle Heuristic (MSCH)}. The FVS is 
obtained by iteratively removing the node that appears on the most number of simple cycles in $G_{dep}$ until no more cycles exist. The simple cycles are 
enumerated.
\item {\bf Maximum Cycle Heuristic (MCH)}. This heuristic is similar to MSCH
but counts cycles differently. For each vertex $v \in V_{dep}$, it finds a 
cycle going through $v$ and marks the outgoing edge from $v$ on this cycle. 
The process is repeated for $v$ until no more cycles can be found. The vertex
with the largest cycle count is then removed first.
\item {\bf Maximum Degree Heuristics (MDH)}. This \changed{heuristic} constructs \changed{an} FVS
through vertex deletion based on the degree of the vertex until no cycles 
exist. 
\end{enumerate}
Based on FVS, the solution minimizing travel distance
can be found by {\sc MinDist}() (line \ref{algo:lwo-model}), which
is an LP modeling method inspired by \cite{YuLaV16} and
described in Appendix C of \cite{HanStiKroBekYu17RSSSup}.

\subsubsection*{Complete algorithm} Note that {\sc ToroFVSSingle}() is a 
complete algorithm for solving  \rwo\ but it is not a complete 
algorithm for solving \rwo\ optimally. With some additional engineering,
a complete optimal \rwo\ solver can also be constructed: under the 
assumption \changed{that grasping dominates} the traveling costs, simply iterate
through all optimal FVS sets and then compute the subsequent minimum 
distance. After all such solutions are obtained, the optimal among these 
are chosen. It turns out that doing this enumeration does not 
provide much gain in solution quality as the optimal distances are very 
similar to each other.



\section{Performance Evaluation}
	\label{sec:performance}
Experiments are executed on a Intel(R) Core(TM) i7-6900K CPU with 32GB RAM at
2133MHz. Concorde \cite{AppBix+07} is used for solving the \tsp and Gurobi
6.5.1 \cite{Gurobi} for ILP models. 

\subsection{TORO-NO: Minimizing the Travel Distance}
\begin{figure}[htp]
		\centering
        \includegraphics[keepaspectratio, width = .77\columnwidth]{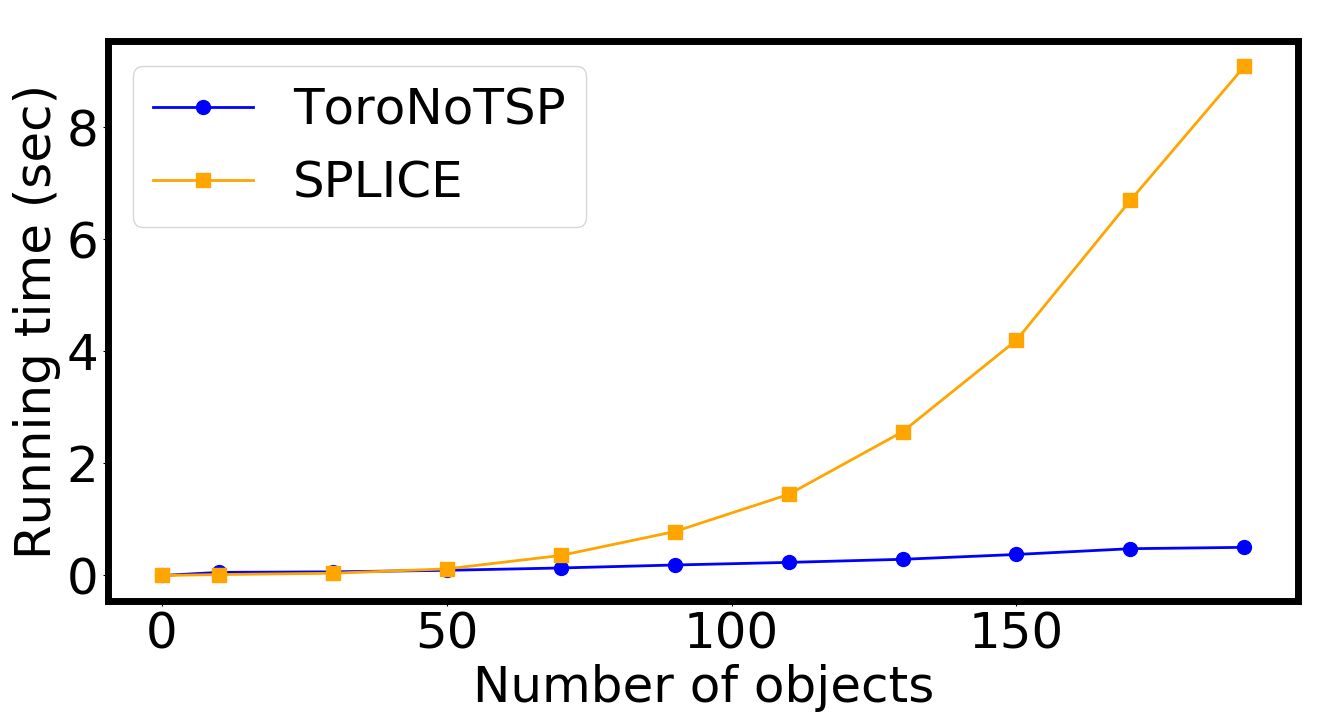}
		\caption{Running time comparison of \rnos and \textsl{SPLICE}.}
		\label{fig:rno-time}
	\end{figure}

\begin{figure}[htp]
		\centering
		\includegraphics[keepaspectratio, width = .77\columnwidth]{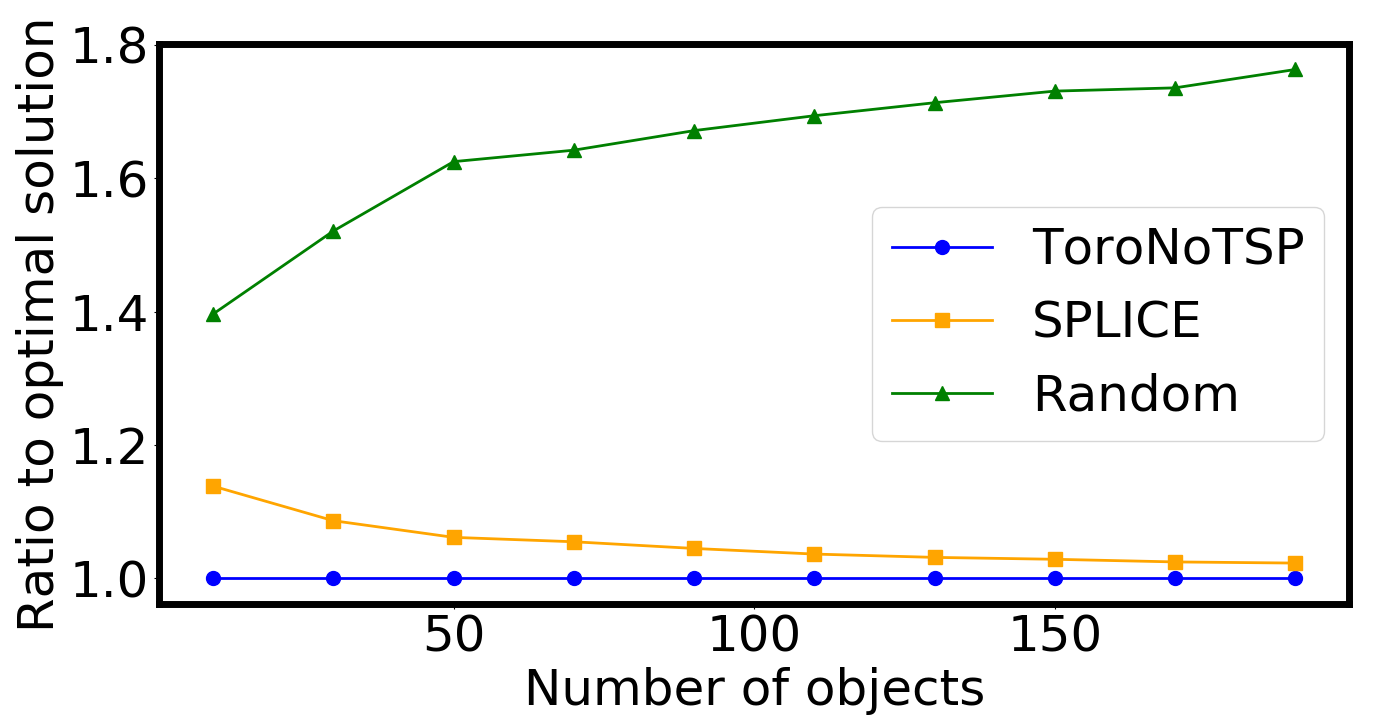}
		\caption{Optimality of \rnos, \textsl{SPLICE} and a random selection method.}
		\label{fig:rno-opt}
	\end{figure}

To evaluate the effectiveness of \rnos, random \rno instances are generated 
in which the number of objects varies. For each choice of number of objects, 100
instances are tried and the average is taken. Although \rnos works on thousands 
of objects (it takes less than $30$ seconds for \rnos to solve instances
with $2500$ objects), the evaluation is limited to $200$
objects\footnote{\changed{State of the art Delta robots have comparable
abilities.  For example, the Kawasaki YF03 Delta Robot is capable of performing
222 pick-and-place actions per minute (1kg objects).}}.
%
Concerning running time, \rnos\ is compared with \textsl{SPLICE} \cite{TrePavFra13} 
which does not compute an exact optimal solution. As shown in 
Fig.~\ref{fig:rno-time}, it takes less than a second for \rnos to compute
the distance optimal manipulator action set. 
Fig.~\ref{fig:rno-opt} illustrates the solution quality of  \rnos, \textsl{SPLICE}, and an algorithm that 
picks a random feasible solution. Notice that the 
random feasible solution generally has poor quality. \textsl{SPLICE} does 
well as the number of objects increases, but under-performs compared to \rnos. 
In conclusion, \rnos provides the best performance on both running time
and optimality for practical sized \rno problems.  

For the unlabeled case (\uno), the same experiments are carried out. The results
appear in Table~\ref{tab:uno}. Note that \textsl{SPLICE} no longer applies.
The last line of the table is the optimality of random solutions, included 
for reference purposes. For larger cases, the \tsp based method is able to 
solve for over $500$ objects in $30$ seconds.
	\begin{table}
		\centering
		\caption{The evaluation of the \tsp model for unlabeled case.}
		\begin{tabular}{l | c | c | c | c}
			\hline
			Number of objects & 10 & 50 & 100 & 200 \\ \hline
			Running time (sec) & 0.04 & 0.58 & 2.43 & 7.30 \\ \hline
			Optimality of random solution & 1.94 & 3.72 & 4.92 & 6.01 \\
			\hline
		\end{tabular}
		\label{tab:uno}
	\end{table}

\subsection{TORO: Minimizing the Number of Grasps}

\begin{figure}[htp]
	    \centering
		\includegraphics[keepaspectratio, width = .77\columnwidth]{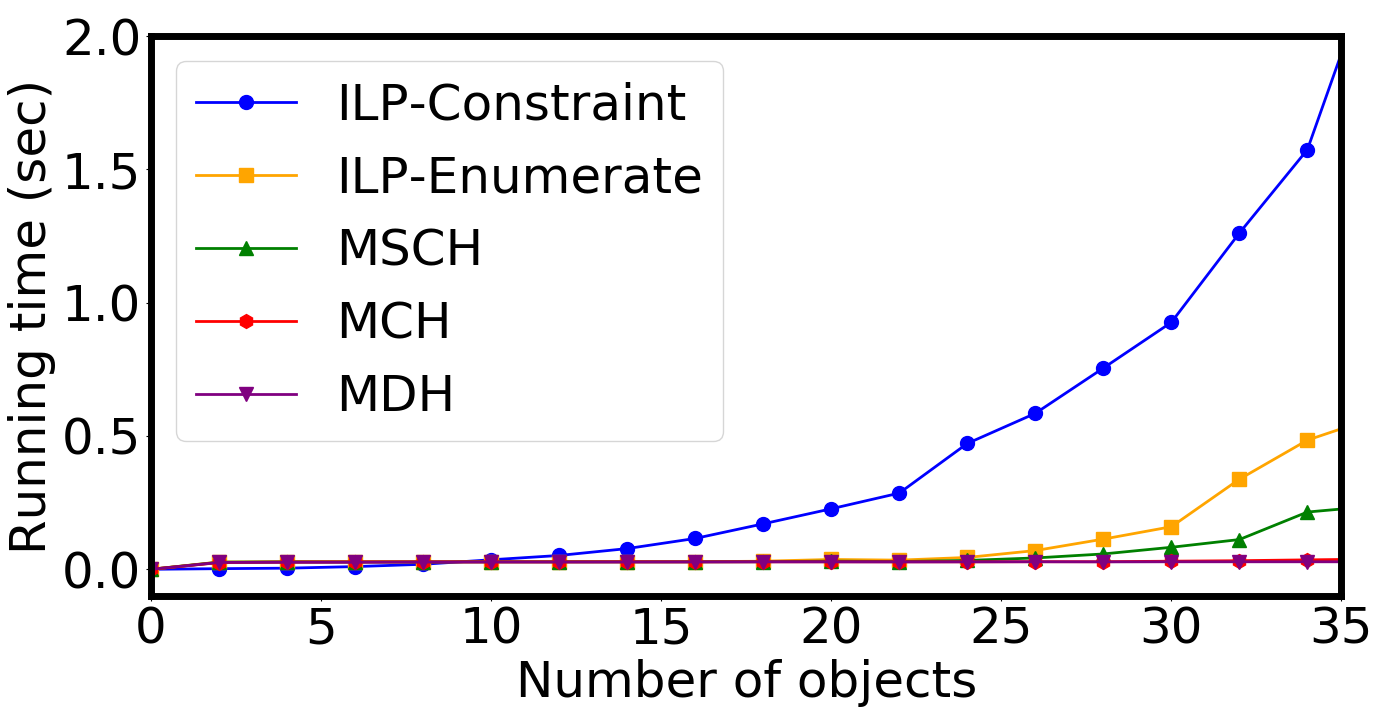}
		\caption{
            Running time of various methods for optimizing FVS. 
        }
		\label{fig:fvs-time}
	\end{figure}

\noindent To evaluate different FVS minimization methods, 
dependency graphs are generated by capping the average degree and maximum
degree for a fixed object count. To evaluate the running time, 
the average degree is set to $2$ and the maximum degree is set to $4$, which
creates significant dependencies
The
running time comparison is given in Fig.~\ref{fig:fvs-time} (averaged
over 100 runs per data point). Although exact ILP-based methods took
more time than heuristics, they can solve optimally for over $30$
objects in just a few seconds, which makes them very practical.

\begin{figure}[htp]
		\centering
		\includegraphics[keepaspectratio, width = .77\columnwidth]{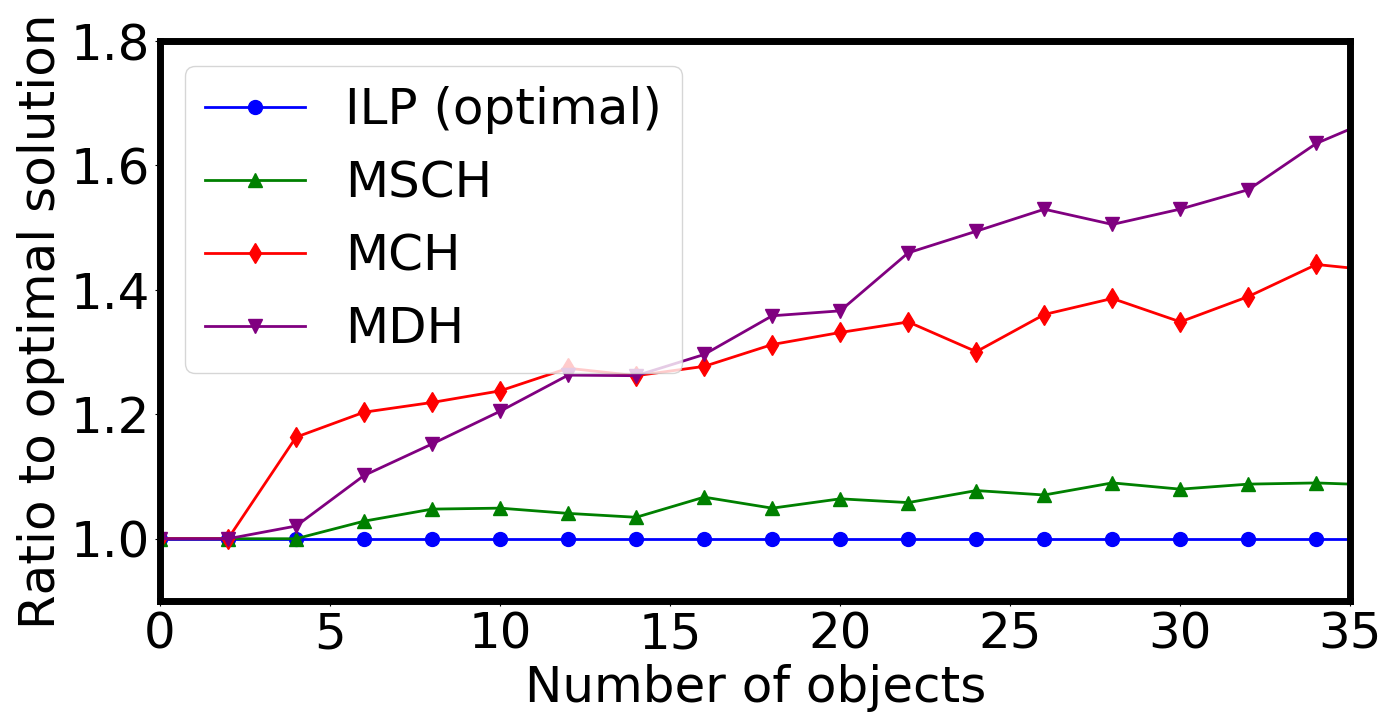}
        \caption{Optimality ratio of various methods for optimizing FVS as 
			compared with the optimal ILP-based methods.}
		\label{fig:fvs-opt}
	\end{figure}

When it comes to performance (Fig.~\ref{fig:fvs-opt}), ILP-based methods
have no competition. Interestingly, simple cycle based method (MSCH) also
works quite well and may be useful in place of ILP-based methods for larger 
problems, given that MSCH runs faster. 
	
\begin{figure}[htp]
		\centering
		\includegraphics[keepaspectratio, width = .77\columnwidth]{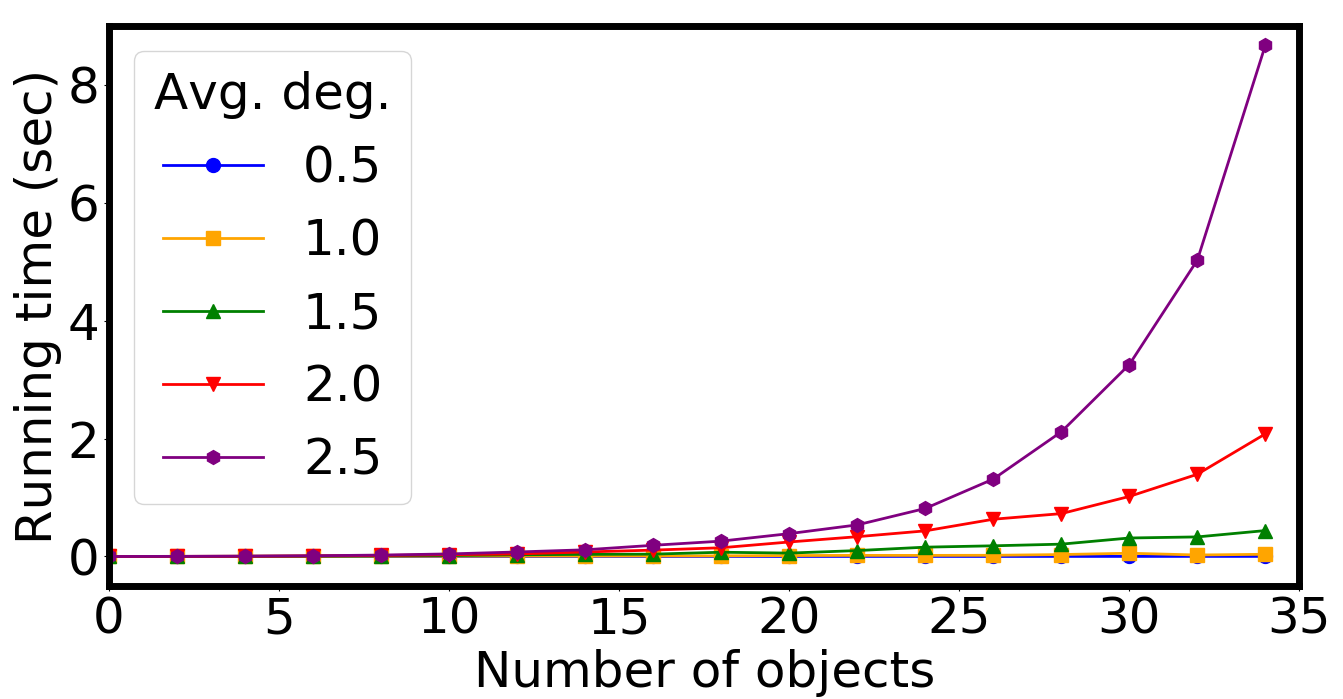}
		\caption{The running time of ILP-Constraint under varying $G_{dep}$
		average degree. Maximum degree is capped at twice the average degree.}
		\label{fig:fvs-dense-eval}
	\end{figure}

The performance is also affected by the average degree for each node, 
which is directly linked to the complexity of $G_{dep}$. Fixating on 
the ILP-Constraint algorithm, average degree of $0.5$-$2.5$ are 
experimented ($2.5$ average degree yields rather constrained dependency 
graphs). As can be observed from Fig.~\ref{fig:fvs-dense-eval}, for
up to $35$ objects, an optimal FVS can be readily computed in a few seconds. 

\begin{figure}[htp]
		\centering
		\includegraphics[keepaspectratio, width = .77\columnwidth]{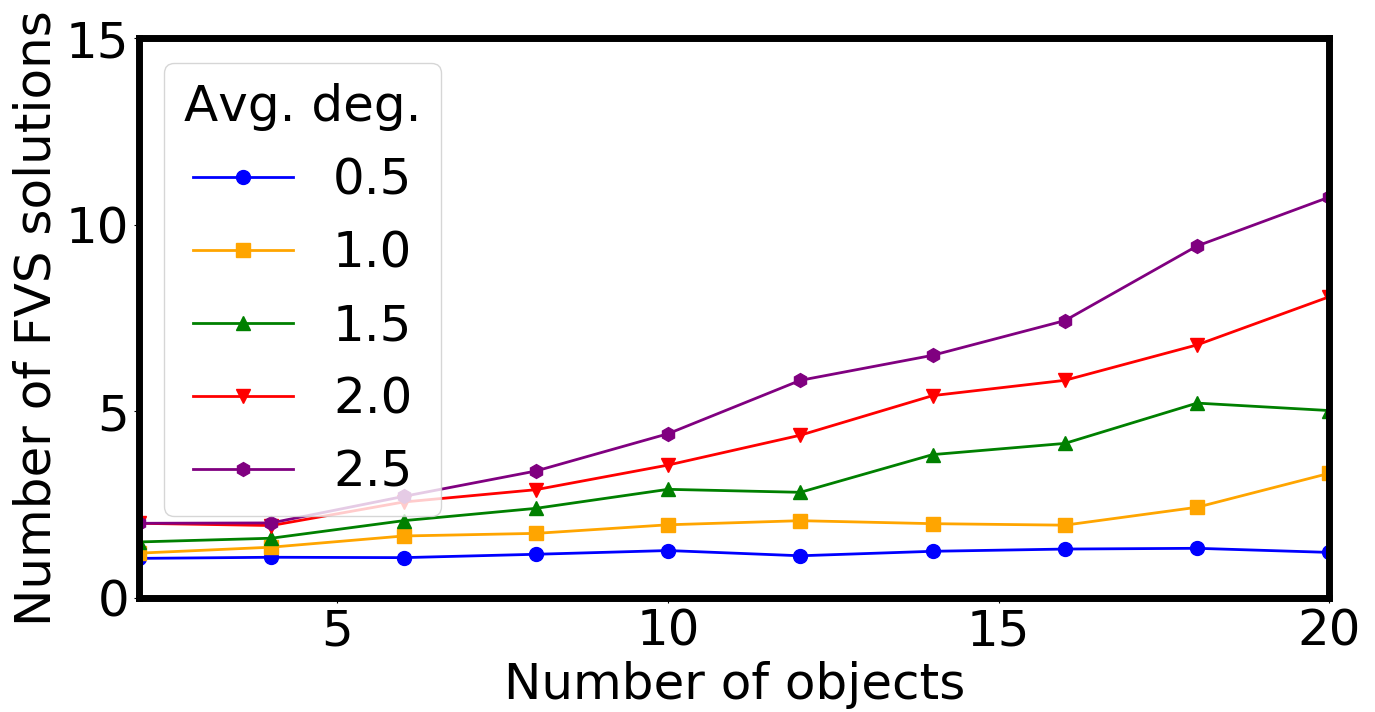}
		\caption{The number of optimal FVS solutions in expectation.}
		\label{fig:fvs-number}
	\end{figure}

Finally, this section emphasizes an observation regarding the number of optimal
FVS sets (Fig.~\ref{fig:fvs-number}). By disabling FVSs that are already
obtained in subsequent runs, all FVSs for a given problem can be exhaustively
enumerated for varying numbers of objects and average degree of $G_{dep}$. The
number of optimal FVSs turns out to be fairly limited.

\subsection{TORO: Overall Performance}
\begin{figure}[htp]
	\centering
	\includegraphics[keepaspectratio, width = .77\columnwidth]{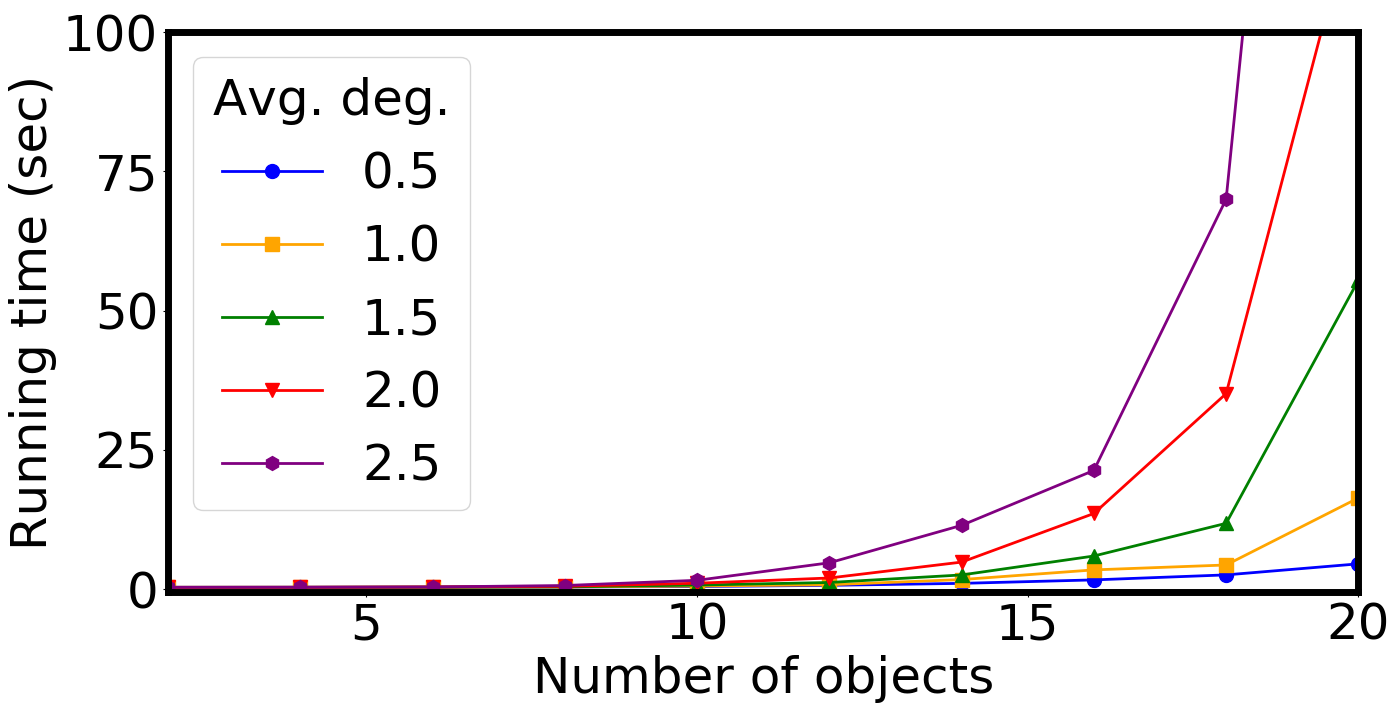}
	\caption{The total running time for {\sc ToroWoFVSSingle}().}
	\label{fig:lwo-total}
\end{figure}

The running time for the entire {\sc ToroWoFVSSingle}() is provided 
in Fig.~\ref{fig:lwo-total}. Observe that FVS computation takes 
almost no time in comparison to the distance minimization step. 
As expected, higher average degrees in $G_{dep}$ \changed{make} the computation 
harder. 

\begin{figure}[htp]
		\centering
		\includegraphics[keepaspectratio, width = .77\columnwidth]{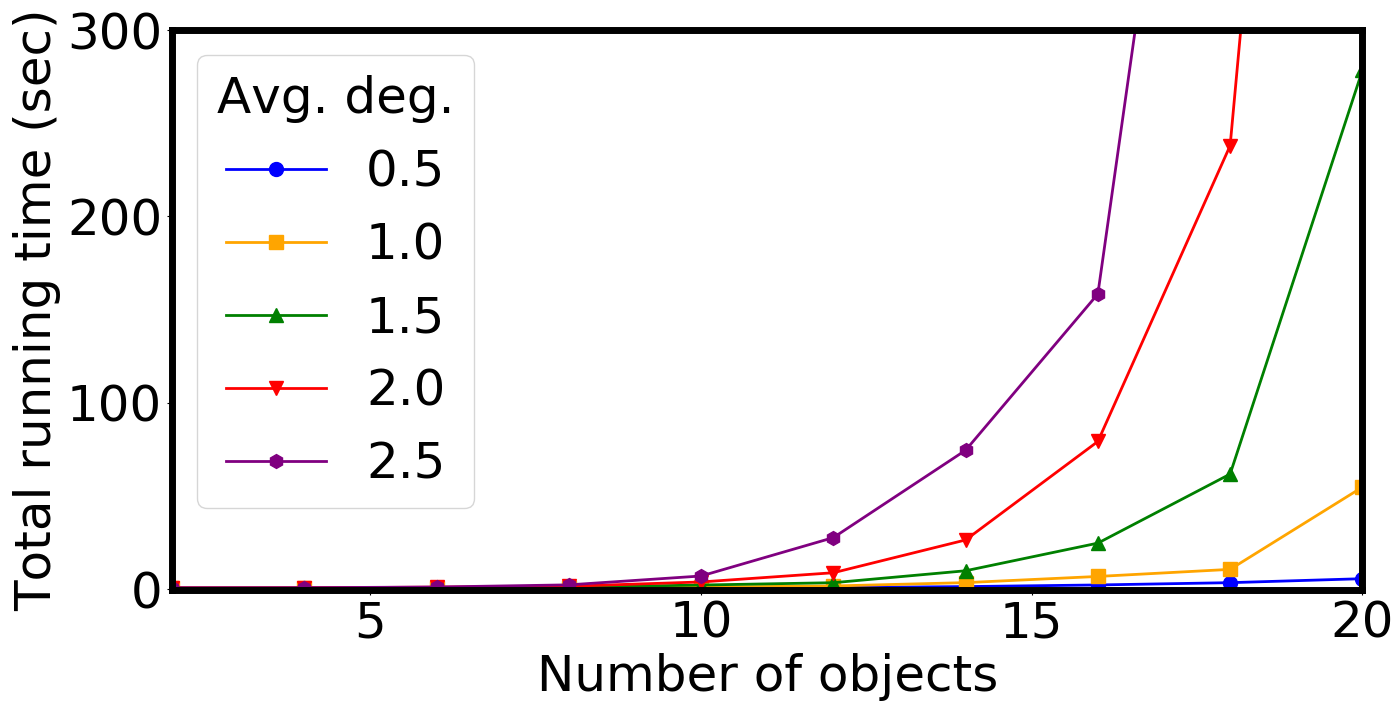}
		\caption{The running time to produce a global optimal solution for \rwo.}
		\label{fig:rwo-global-total}
	\end{figure}

Running {\sc ToroWoFVSSingle}() together with FVS enumeration, 
a global optimal solution is computed for \rwo under the assumption that the 
grasp/release costs dominate. Only solutions 
with an optimal FVS are considered. The computation time is provided in 
Fig.~\ref{fig:rwo-global-total}. The result shows that it gets costly 
to compute the global optimal solution as the number of objects go 
beyond $15$ for dense setups. It is empirically observed that 
for the same problem instance and different optimal FVSs, the minimum 
distance computed by {\sc MinDist}() \changed{in Alg.~\ref{algo:lwo}} has less than $5\%$ variance. 
This suggests that running {\sc ToroWoFVSSingle}() just once should 
yield a solution that is very close to being the global optimal. 

\section{Conclusion} 
	\label{sec:conclusion}
This paper studies the combinatorial structure inherent in tabletop
object rearrangement problems. For \rno and \uno, it is shown that
Euclidean-\tsp can be reduced to them, establishing their
NP-hardness. More importantly, \rno and \uno can be reduced to \tsp
with little overhead, thus establishing that they have similar
computational complexity and lead to an efficient solution
scheme. Similarly, an equivalence was established between dependence
breaking of \rwo and \fvs, which is APX-hard. The equivalence enables
subsequent ILP-based methods for effectively and optimally solving
\rwo instances containing tens of objects with overlapping starts and
goals.
	
   Exploring scenarios in which objects 
	are more tightly entangled, i.e., cases with high object density and thus 
	little ``buffer'' space, remain an open problem. 
   The methods and algorithms in this paper serve as a strong foundation for
   solving complex untangling and rearrangement tasks on tabletops. 
	

\section*{Acknowledgments}
This work is supported by NSF awards IIS-1617744, IIS-1451737 and
CCF-1330789, as well as internal support by Rutgers University.  Any
opinions or findings expressed in this paper do not necessarily
reflect the views of the sponsors. The authors would like to thank the
anonymous RSS reviewers for their constructive comments.

\newpage

\bibliographystyle{IEEEtranN}

\bibliography{bib/nick-ref}    

\appendices
\section{Proof for Cost-optimal \uno}\label{proof-couno}
\begin{proof}[Proof of Theorem IV.2]
Again, reduce from Euclidean-\tsp. The same \tsp instance from the proof of
Theorem~IV.1 is used. The conversion to \rno and the process to
obtain \changed{a} \uno instance are also similar, with the exception being that edges
$s_ig_i$ are not required to be used in a solution; this makes the labeled case
become unlabeled. 

The argument is that the cost-optimal solution of the \uno instance also yields
an optimal solution to the original Euclidean-\tsp tour. This is accomplished
by showing that an optimal solution to the \rno instance has essentially the
same cost as the \uno instance. To see that this is the case, assume that an
optimal solution (tour) path to the reduced \uno problem is given. Let the path
have a total length (cost) of $D_{opt}^{\uno}$. Let $s_ig_i$ be the first such
edge that is not in the \uno solution. Because the path is a tour, following
$s_i$ along the path will eventually reach $g_i$. The resulting path will
have the form $s_iv_1\ldots v_2g_iv_3$, i.e., the black path in
Fig.~\ref{fig:uno-hardness}. 
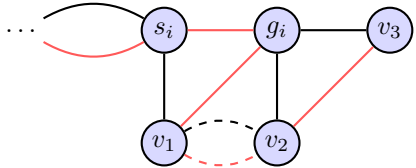
\begin{figure}[htp]
	\begin{center}
	\begin{tikzpicture}[scale=1.5]
		\foreach \nodeName/\nodeLocation in {s_i/{(0, 1)}, v_1/{(0, 0)}, v_2/{(1, 0)}, g_i/{(1, 1)}, v_3/{(2, 1)}}{
			\node (\nodeName) at \nodeLocation {$\nodeName$};
		}
		\node[draw=none, fill=none] (helper) at (-1.25, 1) {$\dots$};
		\foreach \edgeFrom/\edgeTo in {s_i/v_1, g_i/v_2, g_i/v_3}{
			\draw [-] (\edgeFrom) to (\edgeTo);
		}
		\foreach \edgeFrom/\edgeTo in {s_i/g_i, g_i/v_1, v_2/v_3}{
			\draw [-, color=customLightRed] (\edgeFrom) to (\edgeTo);
		}
		\draw [-] (helper) to [out = 30, in = 150] (s_i);
		\draw [-, color=customLightRed] (helper) to [out = -30, in = -150] (s_i);
		\draw [-, dashed] (v_1) to [out = 30, in = 150] (v_2);
		\draw [-, color=customLightRed, dashed] (v_1) to [out = -30, in = -150] (v_2);
	\end{tikzpicture}
	\end{center}
	\caption{\label{fig:uno-hardness} Augmenting a path in an \uno
solution.}  
\end{figure}

Upon the observation of such a partial solution, proceed to make 
the augmentation and replace the path with the new one (red path in 
Fig.~\ref{fig:uno-hardness}). Because $s_ig_i \ll 1/(4n)$, the potential 
increase in path length is bounded by (note that $v_2v_3$ is shorter
than the additive length of $v_2g_i$ and $g_iv_3$)
\[
\|s_ig_i\|_2 + \|g_iv_1\|_2 - \|s_iv_1\|_2 \le 2\varepsilon \ll 1/(2n). 
\]

After at most $n$ such augmentations, an optimal \uno solution is 
converted to an \rno solution. The \rno solution has a cost increase of 
at most $n*1/(2n) = 1/2$. The \rno solution can then be converted
to a solution of the Euclidean-\tsp problem, which will not increase the 
cost. Thus, a \uno solution can be converted to a corresponding 
Euclidean-\tsp solution with a cost addition of less than $1/2$. Let the 
Euclidean-\tsp solution obtained in this manner have a total cost of $D'$, then 
\begin{equation}\label{equ:tuno1}
D' < D^{\uno}_{opt} + \frac{1}{2}.
\end{equation}

Now again let the optimal Euclidean-\tsp solution have a cost of $D_{opt}$.
The solution can be converted to an \rno solution with a total cost
of less than $D_{opt} + 1/4$. The \rno solution is also a solution to 
the \uno problem. That is, for this new \uno solution, the cost is 
\begin{equation}\label{equ:tuno2}
D^{\uno} < D_{opt} + \frac{1}{4}. 
\end{equation}
Now, if $D' > D_{opt}$, then $D' \ge D_{opt} + 1$. Putting this 
together with~\eqref{equ:tuno1} and~\eqref{equ:tuno2}, 
\[
D^{\uno} < D_{opt} + \frac{1}{4} \le D' - \frac{3}{4} \le D^{\uno}_{opt} - \frac{1}{4},
\]
which is a contradiction. Therefore, $D' > D_{opt}$ cannot be true.
Therefore, a cost-optimal \uno solution yields an optimal solution to 
the original Euclidean-\tsp problem. This shows that \uno is at least as 
hard as Euclidean-\tsp.
\end{proof}
\section{Exact ILP-Based Algorithms for Finding Optimal FVS}
To compute the exact solution, the problem is modeled as an ILP problem, and 
then solved using LP solvers, e.g., Gurobi \tsp Solver \cite{Gurobi}. In this 
paper two different ILP models is used, which are similar to the models 
introduced in section 3.1 and 3.2 of \cite{BahSchNeu15}:
\begin{enumerate}
  \item \textbf{ILP-Constraint.} By splitting all vertices $o_i \in G_{dep}$ to $o_i^{in}$ and $o_i^{out}$, a new graph $G_{arc}(V_{arc}, E_{arc})$ is constructed, where $V_{arc} = \{o_1^{in}, o_1^{out}, \dots, o_n^{in}, o_n^{out}\}$, and $(o_i^{out}, o_j^{in}) \in E_{arc}$ \textit{iff} $(o_i, o_j) \in A_{dep}$. By adding extra edges $(o_i^{in}, o_i^{out})$ for all $1 \leq i \leq n$ to $E_{arc}$, problem is transformed to a \textsl{minimum feedback arc set} problem, where the objective is to find a minimum set of arcs to make $G_{arc}$ acyclic. Moreover, every edge in this set ends at $o_i^{in}$ or starts from $o_i^{out}$ can be replaced by $(o_i^{in}, o_i^{out})$, which stands for a vertex $o_i$ in $G_{dep}$, without changing the feasibility of the solution. The proof is omitted due to the lack of space.
		
      The next step is to find a minimum cost ordering $\pi^*$ of the nodes in $G_{arc}$.
      Let $c_{i,j} = 1$ if edge $(i,j) \in E_{arc}$, while $c_{i,j} = 0$ if
      edge $(i,j) \notin E_{arc}$. Furthermore, let binary variables $y_{i,j}$
      associate the ordering of $i,j \in \pi$, where $y_{i,j} = 0$ if $i$
      precedes $j$, or $1$ if $j$ precedes $i$. Suppose $|V_{arc}| = m$, the LP formulation is expressed as:
		\begin{equation*}
			\begin{array}{lrl}
			\displaystyle \min_{y} & \multicolumn{2}{l}{\displaystyle \sum_{j = 1}^m (\displaystyle \sum_{k = 1}^{j - 1} c_{k,j} y_{k,j} + \displaystyle \sum_{l = j + 1}^n c_{l,j}(1 - y_{j,l}))} \\
			\textrm{s.t.} &y_{i,j} + y_{j,k} - y_{i,k} \leq 1,&  1 \leq i < j < k \leq m \\
			&- y_{i,j} - y_{j,k} + y_{i,k} \leq 0,&  1 \leq i < j < k \leq m \\
			\end{array}
		\end{equation*}
		
		The solution arc set contains all the backward edges in $\pi^*$.

		\item \textbf{ILP-Enumerate.} First find the set $C$ of all the simple cycles in $G_{dep}$. A set of binary variables $V = \{v_1, \dots, v_n\}$ is defined, each assigned to an object $o_i \in O$, the LP formulation is expressed as::
		\begin{equation*}
			\begin{array}{lll}
			\displaystyle \max_{v} & \multicolumn{2}{l}{\displaystyle \sum_{v_i \in V} v_i} \\
			\textrm{s.t.} & \displaystyle \sum_{o_i \in C_j} v_i < |C_j|, &\forall C_j \in C. \\
			\end{array}
		\end{equation*}		

		Then the vertices in the minimum FVS is the objects whose corresponding variable $v_i$ is 0 in the solution of this LP model.
	\end{enumerate}

\section{Algorithm for Shortest Travel Distance}\label{sec:lwolp}
	
   Given $n$ objects $\objects = \{o_1, \dots, o_n\}$, the minimum FVS $B =
   \{o_1, \dots, o_p\}$. The solution that yields the shortest travel distance has constraints on FVS set $B$ and $n + p$ grasps. Meanwhile, the upper
   bound on the number of buffers is $|B| = p$.
	
   The total travel distance is minimized based on $n + p$ actions. For each
   step Boolean variables are introduced in order to denote the location of the
   variables. At time step $t$, the start and goal locations are $S^t =
   \{s_1^t, \dots, s_n^t\}$ and $G^t = \{g_1^t, \dots, g_n^t\}$, where each
   element defines whether the object is at this position. Buffers are
   represented as copies, which is $B^t = \{b_{11}^t, b_{12}^t,
   \dots,b_{pp}^t\}$, where $b_{ij}^t = 1$ indicates $o_i$ is in buffer $b_j$
   at time step $t$. Then for $1 \leq i \leq n, 1 \leq j \leq p, 1 \leq k \leq
   p$, the initial state is
	\[s_i^0 = 1, g_i^0 = 0, b_{jk}^0 = 0.\]
	The goal state after rearrangement is
	\[s_i^{n + p} = 0, g_i^{n + p} = 1, b_{jk}^{n + p} = 0.\]
	
	Boolean variables are linked to edges between two time steps. If $e(g_i^t s_j^{t + 1}) = 1$ denotes moving the manipulator from $g_i$ to $s_i$ between time step $t$ and $t + 1$, then between time step 0 and 1: 
	\[\sum_{i = 1}^n e(s_M s_i^1) = 1, s_i^1 = s_i^0 - e(s_M s_i^1).\]
	When $1 \leq i \leq p$,
	\[\sum_{j = 1}^p e(s_i^1 b_{ij}^1) = e(s_M s_i^1), b_{ij}^1 = b_{ij}^0 + e(s_i^1 b_{ij}^1).\]
	When $p + 1 \leq i \leq n$,
	\[e(s_i^1 g_i^1) = e(s_M s_i^1), g_i^1 = g_i^0 + e(s_i^1 g_i^1).\]
	
	For time step $1 \leq t \leq n + p - 1$:
	\begin{enumerate}
		\item 
			For the edges going out of this time step, when $1 \leq i \leq p$, $1 \leq j \leq p$,
            {\small\[e(s_i^t b_{ij}^t) = \sum_{k = 1}^m e(b_{ij}^t s_k^{t + 1}) + \sum_{k = 1}^p \sum_{l = 1}^p e(b_{ij}^t b_{kl}^{t+1}).\]}
			When $1 \leq i \leq p$,
            {\small\[\sum_{j = 1}^{p} e(b_{ij}^t g_i^t) = \sum_{k = 1}^n e(g_i^t s_k^{t + 1}) + \sum_{k = 1}^p \sum_{l = 1}^p e(g_i^t b_{kl}^{t+1}).\]}
			When $p + 1 \leq i \leq n$,
            {\small\[e(s_i^t g_i^t) = \sum_{k = 1}^n e(g_i^t s_k^{t + 1}) + \sum_{k = 1}^p \sum_{l = 1}^p e(g_i^t b_{kl}^{t+1}).\]}
			
		\item
			Then for time step $t+1$ constraint are imposed for the incoming edges, in order to avoid the scenario where the manipulator travels to an empty location. For $1 \leq i \leq n$,
            {\small\[\sum_{j = 1}^n e(g_j^t s_i^{t+1}) + \sum_{j = 1}^p \sum_{k = 1}^p e(b_{jk}^t s_i^{t+1}) \leq s_i^t.\]}
			For $1 \leq i \leq p$, $1 \leq j \leq p$,
            {\small\[\sum_{k = 1}^n e(g_k^t b_{ij}^{t+1}) + \sum_{k = 1}^p \sum_{l = 1}^p e(b_{kl}^t b_{ij}^{t+1}) \leq b_{ij}^t.\]}
	
		\item
			Then for the edges inside step $t + 1$, when $1 \leq i \leq p$,
            {\small \[\sum_{j = 1}^{p} \sum_{k = 1}^p e(s_i^{t+1} b_{jk}^{t+1}) = \sum_{j = 1}^n e(g_j^t s_i^{t+1}) + \sum_{j = 1}^p \sum_{k = 1}^p e(b_{jk}^t s_i^{t+1}).\]}
			When $p + 1 \leq i \leq n$,
            {\small\[e(s_i^{t+1} g_i^{t+1}) = \sum_{j = 1}^n e(g_j^t s_i^{t+1}) + \sum_{j = 1}^p \sum_{k = 1}^p e(b_{jk}^t s_i^{t+1}).\]}
			And for all $1 \leq i \leq p$, $1 \leq j \leq p$,
            {\small\[e(b_{ij}^{t+1} g_i^{t+1}) = \sum_{k = 1}^n e(g_k^t b_{ij}^{t+1}) + \sum_{k = 1}^p \sum_{l = 1}^p e(b_{kl}^t b_{ij}^{t+1}).\]}

		\item 
			After setup the edges, the variables in step $t+1$ are updated, for $1 \leq i \leq n$,
            {\small\[s_i^{t + 1} = s_i^t - \sum_{j = 1}^n e(g_j^t s_i^{t+1}) - \sum_{j = 1}^p \sum_{k = 1}^p e(b_{jk}^t s_i^{t+1}).\]}
			And for $1 \leq i \leq p$, $1 \leq j \leq p$,
            {\small \[b_{ij}^{t+1} = b_{ij}^t + e(s_i^{t+1} b_{ij}^{t+1}) - \sum_{k = 1}^n e(g_k^t b_{ij}^{t+1}) - \sum_{k = 1}^p \sum_{l = 1}^p e(b_{kl}^t b_{ij}^{t+1}).\]}
            {\small\[g_i^{t + 1} = g_i^t + \sum_{j = 1}^p  e(b_{ij}^{t+1} g_i^{t+1}).\]}
			And for $p + 1 \leq i \leq n$,
            {\small\[g_i^{t + 1} = g_i^t + e(s_i^{t + 1} g_i^{t+1}).\]}
			
		\item 
			Since the buffers are presented in multiple copies, constraint must be designed to make each buffer is occupied by at most one object, so for $1 \leq i \leq p$,
            {\small\[\sum_{j = 1}^p b_{ji}^t \leq 1\]}
			
		\item 
			Then the constraints for dependencies. Suppose $s_i$ is in collision with $g_j$ then
            {\small\[s_i^t + g_j^t \leq 1\]}
	\end{enumerate}
	
	Finally, the manipulator goes to $g_M$. So for $1 \leq i \leq p$,
    {\small\[e(g_i^{n + p} g_M) = \sum_{j = 1}^p \sum_{k = 1}^p e(b_{jk}^{n + p} g_i^{n + p})\]}
	And for $p + 1 \leq i \leq n$,
    {\small\[e(g_i^{n + p} g_M) = e(s_i^{n + p} g_i^{n + p})\]}
	
	The cost function of this LP:
    {\small\[\min \sum_{e(ab) \text{ in model}} \text{dist}(a, b).\]}


\end{document}